\documentclass[letterpaper]{article} 
\usepackage[preprint]{aaai2027}  
\usepackage[hyphens]{url}  
\usepackage{graphicx} 
\usepackage{natbib}  
\usepackage{caption} 
\usepackage{algorithm}
\usepackage{algorithmic}

\usepackage{newfloat}
\usepackage{listings}
\DeclareCaptionStyle{ruled}{labelfont=normalfont,labelsep=colon,strut=off} 
\floatstyle{ruled}
\newfloat{listing}{tb}{lst}{}
\floatname{listing}{Listing}

\usepackage{booktabs}

\usepackage{color}

\definecolor{cb_orange}{rgb}{1.0,0.51,0.0}
\definecolor{cb_blue}{rgb}{0.22,0.49,0.72}
\definecolor{cb_green}{rgb}{0.3,0.67,0.29}
\definecolor{cb_red}{rgb}{0.89,0.1,0.11}
\definecolor{cb_pink}{rgb}{1, 0, 0.4}
\definecolor{cb_purple}{rgb}{0.5,0.0,0.5}
\definecolor{cb_cyan}{rgb}{0.0,1.0,1.0}
\definecolor{cb_magenta}{rgb}{1.0,0.0,1.0}

\usepackage{makecell}
\usepackage{multirow}
\usepackage{booktabs}
\usepackage{array} 
\usepackage{graphicx}
\usepackage{caption}
\usepackage{pifont}
\newcommand{\xmark}{\ding{55}}
\usepackage{newunicodechar}
\newunicodechar{−}{\textminus}
\usepackage{newfloat}
\usepackage{listings}
\usepackage{amsmath}
\usepackage{amssymb}
\usepackage{enumitem}
\usepackage{tabularx} 
\usepackage{colortbl} 

\title{LanGuSTE: Language-Guided Coarse-to-Fine Patch Selection \\ for Efficient Whole Slide Image Analysis}

\author {
    Yonghan Shin,
    Gangsu Kim,
    Won-Ki Jeong$^\dagger$
}
\affiliations {
    Department of Computer Science and Engineering, College of Informatics,
    Korea University, Republic of Korea\\
    \{dydgks592,gangsu1813,wkjeong\}@korea.ac.kr
}

\begin{document}

\maketitle

\def\thefootnote{$\dagger$}\footnotetext{Corresponding author.}

\begin{abstract}
Whole slide images (WSIs) in computational pathology pose a major computational challenge due to their gigapixel scale, often requiring tens to hundreds of thousands of high-resolution patches to be processed per slide. 
%
In conventional WSI pipelines, exhaustive high-resolution patch processing makes preprocessing far more time-consuming than downstream model training. Existing patch selection methods suffer from a fundamental paradox: all patches must still be extracted and encoded at least during training, and sometimes during both training and inference, before irrelevant ones can be discarded.
To address this, we propose LanGuSTE, an efficient patch selection framework that integrates pathology-domain vision-language models (VLMs) and knowledge derived from large language models (LLMs) through two key modules: Cross-Scale Visual Prompt Tuning (CS-VPT) and coarse-to-fine patch selection. CS-VPT aligns low-resolution patches with their spatially corresponding high-resolution patches through contrastive learning, transferring fine-grained diagnostic semantics into low-resolution representations. The patch selection module then leverages VLM representations and LLM-generated pathology-specific descriptions to identify informative regions in a coarse-to-fine manner, encoding only the corresponding high-resolution patches to reduce preprocessing time. 
Extensive experiments demonstrate that LanGuSTE reduces overall WSI processing time to approximately 3$\times$ while achieving diagnostic performance comparable to or better than exhaustive patch processing and recent state-of-the-art patch-selection methods.
\end{abstract}


\section{Introduction}
\begin{figure}[t]
\centering
\includegraphics[width=0.85\linewidth]{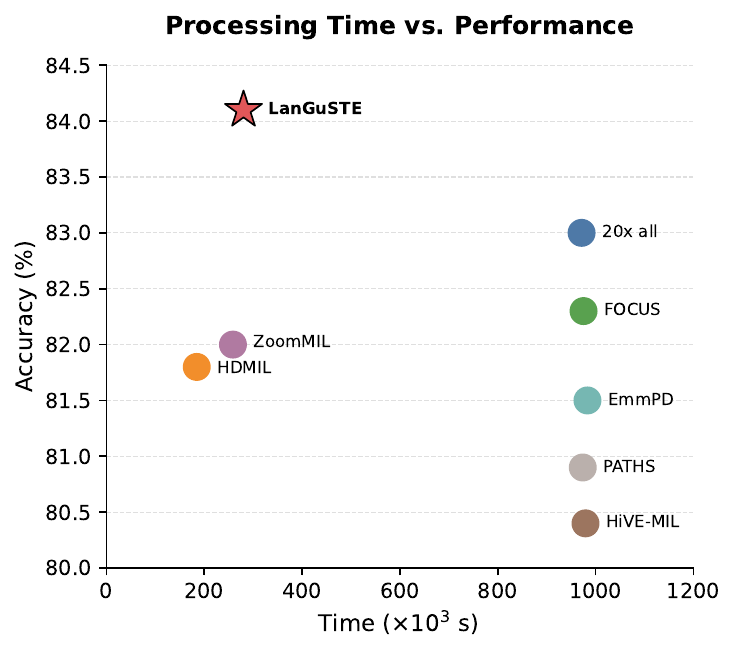}
\caption{Comparison of the proposed method with state-of-the-art methods in terms of average classification accuracy (“Avg.” column of Table~\ref{tab:table_main}) and WSI processing time across six tasks.
%
This plot demonstrates that LanGuSTE achieves higher classification accuracy with shorter processing times than state-of-the-art methods.
}
\label{fig:scatter}
\end{figure}

With recent advances in digital imaging and artificial intelligence~(AI), analyzing pathological data using computational methods, i.e., computational pathology~(CPath), plays a vital role in accurate disease diagnosis and treatment planning.
For the effective implementation of CPath, high-resolution (HR) whole slide images~(WSIs) are essential, as they provide a hierarchical representation of tissue at multiple magnifications, where lower magnifications (e.g., 5$\times$) capture tissue-level architecture and higher magnifications (e.g., 20$\times$) reveal detailed cellular morphology.

Owing to the gigapixel scale of WSIs, it is standard practice to partition each WSI into thousands to hundreds of thousands of small image patches, and to employ a multiple instance learning (MIL) framework in which each patch is treated as an individual instance, and the entire slide is considered a bag~\cite{abmil}. 
However, incorporating the entire set of patches in the MIL framework is not only computationally intensive but also suboptimal for performance. 
~\citet{hdmil} observed that the majority of patches contribute minimally to the final prediction and may even introduce noise. 
Furthermore, we found that a substantial portion of computational time is consumed by patch-level image encoding in the MIL framework.
For instance, encoding all patches from the TCGA-NSCLC dataset~(3,129 WSIs, total size 1,150 GB)~\cite{tcgareport} required approximately two weeks while training a model took less than an hour on an RTX 4090 GPU, making patch encoding a significant bottleneck in building AI models for CPath.

Several prior studies have sought to reduce computational cost in WSI analysis by employing region-of-interest (ROI) detection and patch selection strategies.
%
Early work~\cite{rcnnroi, transformerroi} introduced using a pre-trained ROI detector, which requires supervised training using substantial manual annotations. 
Recent methods such as ZoomMIL~\cite{zoommil} and HDMIL~\cite{hdmil} aim to reduce inference-time computation by selectively analyzing task-relevant patches. 
While effective at filtering irrelevant regions, these approaches still require extracting and encoding all HR patches, resulting in substantial computational and memory costs that limit scalability. 
This creates a paradox: the entire workflow—and often model training itself—must process every patch before determining which ones can be discarded. 
Consequently, the ~\textbf{true bottleneck} in computational pathology remains unresolved.

To address these limitations, we propose~\textbf{LanGuSTE}, a method that selects diagnostically relevant patches in a coarse-to-fine manner without the need to exhaustively encode all HR patches or train a separate ROI selector.
 The key idea is to leverage recent vision-language models (VLMs) tailored to medical and pathology domains, which provide rich textual descriptions aligned with visual features in pathology images.
 By filtering low-resolution (LR) patches based on class-specific morphological descriptions using a pathology VLM in a zero-shot fashion, large numbers of diagnostically irrelevant patches can be discarded early, allowing computational resources to focus on diagnostically meaningful regions.
However, a critical challenge in this approach is that LR images lack the fine-grained visual cues—such as cellular-level morphological features—that are essential for accurate diagnosis. 
%
%
We address this by introducing
Cross-Scale Visual Prompt Tuning (CS-VPT), which aligns LR patches with their spatially corresponding HR counterparts and separates them from unrelated regions, ensuring that subtle diagnostic features are fused in 
%
LR patch features.
%
Additionally, we incorporate large language models (LLMs), such as GPT~\cite{gpt5} and Gemini~\cite{gemini}, to enable efficient patch-level label assignment and to strengthen the selection of diagnostically important visual features.
%
%
The main contributions are summarized as follows:
\begin{center}
\begin{itemize}
\item[$\bullet$] We propose LanGuSTE, a novel approach in CPath that leverages VLMs for efficient patch selection and encoding for WSIs. 
%
The method can quickly identify diagnostically relevant patches in a coarse-to-fine manner without preprocessing all HR patches across the dataset.

\item[$\bullet$] We introduce CS-VPT, which aligns HR information into LR representations via learnable prompts and contrastive loss trained on a few representative slides. 
This enables effective selection of salient patches at a coarse level while preserving fine-grained diagnostic features.


\item[$\bullet$] Through extensive experiments across multiple datasets, we demonstrate that LanGuSTE achieves an approximately threefold reduction in WSI processing time while outperforming both full-patch processing and other competing methods~(Figure~\ref{fig:scatter}).


\end{itemize}
\end{center}

\section{Related Work}
\subsection{ROI Detection \& Patch Selection in CPath}

To address the high computational cost of processing gigapixel WSIs, numerous approaches have been proposed to identify diagnostically relevant regions. 
Early methods focused on ROI detection, using low-level features and superpixel segmentation, convolutional neural networks, or vision transformers to automate ROI extraction~\cite{fastroi,rcnnroi,transformerroi}.
ROI-centric analysis was also shown to improve classification performance~\cite{benefitsroi}.
However, these methods 
mostly \textbf{rely on extensive annotations or pretrained ROI detectors}, which are costly and limit their scalability across tasks or datasets. 

To mitigate the computational burden during inference, patch selection strategies have also been explored. ZoomMIL~\cite{zoommil} and PATHS~\cite{paths} mimic the diagnostic workflow of pathologists by progressively zooming in from LR thumbnails to selectively analyze HR regions.
HiVE-MIL~\cite{hivemil} utilizes a pretrained VLM to compute similarities between LR patches and class-specific prompts, masking irrelevant regions and propagating relevance to HR counterparts.
HDMIL~\cite{hdmil} adopts a differentiable binarization scheme using the Gumbel-sigmoid trick to select task-relevant patches while maintaining consistency with a full-resolution teacher.
FOCUS~\cite{focus} removes redundant patches via a sliding window mechanism and aggregates similar ones using language-guided token selection and pairwise similarity.
EmmPD~\cite{emmpd} similarly applies a sliding-window strategy to perform 2D visual compression and selects diagnostically important patches based on attention scores.
While these patch selection methods reduce inference-time computation by filtering out irrelevant regions, they share a key limitation with ROI detection approaches: \textbf{during training, all HR patches still need to be extracted and encoded,} leading to high computational and memory costs that undermine their overall scalability.

\begin{figure*}[t]
\centering
\includegraphics[width=1\linewidth]{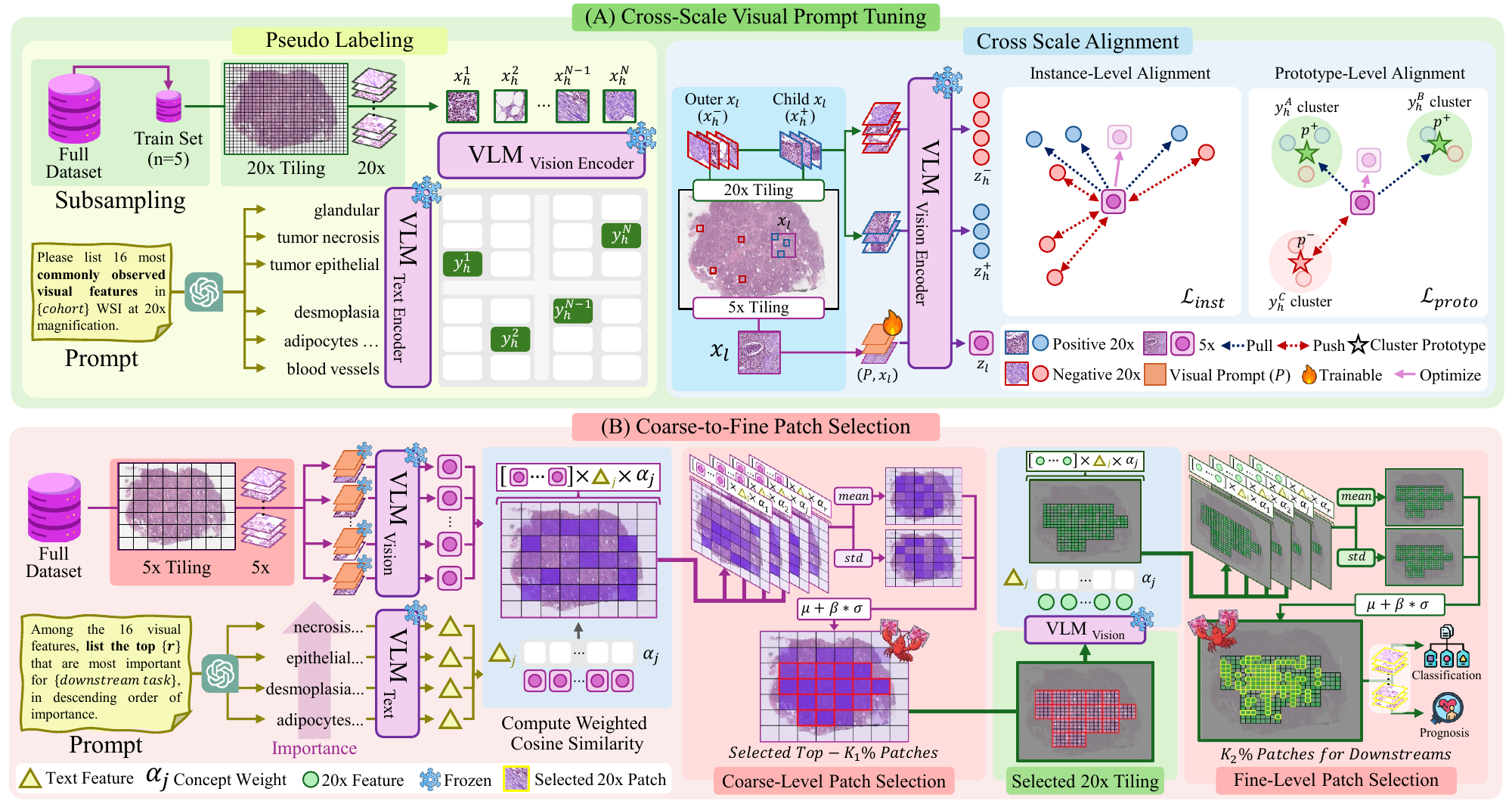}
\caption{Overview of our \textbf{LanGuSTE} framework.
(A) Cross-scale visual prompt tuning (CS-VPT) transfers HR semantic information to the corresponding LR representations using representative WSIs from each class.
(B) Informative patches are selected based on the similarity between the enhanced LR representations and task-relevant pathological concept descriptions.}
\label{framework}
\end{figure*}

\subsection{Vision-Language Models in CPath}
Recent advancements in VLMs and foundation models have substantially improved performance in natural image and text tasks. CLIP~\cite{clip} aligned image and text representations via contrastive learning, effectively bridging the gap between modalities.
%
Inspired by this, researchers have extended VLMs to the pathology domain.
PLIP~\cite{plip} curated domain-specific image–text pairs from public forums to establish a specialized foundation model for pathology.
In parallel, CONCH~\cite{conch} extended this line of work by scaling up the available datasets and broadening the scope of downstream tasks to include not only classification but also generative tasks such as image captioning.
QuiltNet~\cite{quiltnet} continued this trajectory by constructing QUILT-1M from expert-narrated videos and achieving strong performance across classification and retrieval tasks. 
More recently, Patho-CLIP~\cite{pathor1} additionally mined authoritative pathology textbooks to construct a 3.5-million image–text corpus, reporting improved zero-shot classification and cross-modal retrieval over prior pathology VLMs.
%
Although these pathology-specific VLMs have substantially advanced multimodal understanding in digital pathology, they primarily focus on representation learning and downstream task performance.
%
%
\textbf{To the best of our knowledge, no prior work has proposed a principled framework that leverages VLMs to enable time-efficient gigapixel WSI processing by directly addressing the preprocessing bottleneck inherent in existing workflows.} 
Our work represents an initial step in this direction.

\section{Method}

\subsection{Overview}
LanGuSTE 
fully exploits the strengths of a pathology-specialized VLM and an LLM to selectively identify a diagnostically critical subset of patches in WSIs, as illustrated in Figure~\ref{framework}. 
We select the most informative areas in the WSI based on LR representations by computing the similarity between visual and text embeddings from a pre-trained VLM. 
%
To mitigate the risk of losing fine-grained information when using LR patches, we first select a small set of representative WSIs for each task and apply visual prompt tuning~\cite{vpt} to align each LR patch with its corresponding child HR patches and their associated cluster prototypes. 
This alignment is guided by a VLM pretrained on HR pathology images to facilitate effective patch selection.
Then, we conduct a top-k selection based on the similarity between these enhanced LR embeddings and the text embeddings, and subsequently encode the corresponding HR patches, thereby filtering out irrelevant regions and improving efficiency by alleviating the preprocessing bottleneck.

\subsection{Contrastive Alignment via CS-VPT}
To indirectly guide LR patches with the knowledge of their HR counterparts, we perform a contrastive learning--based visual prompt tuning, called CS-VPT.
%
Because the pathology-specific VLM is pretrained on large-scale histopathology corpora and already encodes rich domain knowledge, CS-VPT requires only a small subset of $n$ slides per class; in our experiments, we set $n=5$.
If pathology reports are available, train slides are chosen based on simple cosine similarity between report embeddings (TCGA in Table~\ref{tab:table_main}); otherwise, slides are randomly sampled (UBC-OCEAN and BRACS in Table~\ref{tab:table_main}).
%
We extract HR patches at 20$\times$ magnification and LR patches at 5$\times$ to capture fine-grained and coarse-level tissue features, respectively. 
We then obtain HR patch embeddings using the vision encoder $f_v(\cdot)$ of a pathology-domain VLM.

\paragraph{High-resolution Pseudo-labeling}
To obtain pseudo-labels for HR patches, we leverage the knowledge of LLMs.
For each dataset, we prompt the LLM with a pathology-aware query:
\textit{``Please list $K{=}16$ most commonly observed visual features in \texttt{\{cohort\}} WSIs at 20$\times$ magnification.''}
This yields a set of 16 dataset-specific morphological pseudo-labels representing diagnostically salient visual patterns.
Each pseudo-label is encoded via the VLM’s text encoder $f_t(\cdot)$ with the same text ensemble used in CONCH~\cite{conch} (e.g., "a photo of tumor cells" and "an H\&E stained image of tumor cells"), producing textual embeddings $\{\mathbf{t}_k\}_{k=1}^{K}$.
Given an HR patch $\mathbf{x}_h$, we obtain its visual embedding $\mathbf{z}_h = f_v(\mathbf{x}_h)$.
The pseudo-label assignment is determined by the maximum cosine similarity:
\begin{equation}
    y_h = \arg\max_{k} \frac{\mathbf{z}_h \cdot \mathbf{t}_k}{\|\mathbf{z}_h\| \, \|\mathbf{t}_k\|}.
\end{equation}

\paragraph{Cross-scale Contrastive Prompt Alignment}
A learnable visual prompt embedding $\mathbf{P}$ is passed through the same vision encoder together with each LR patch $\mathbf{x}_l$, yielding LR embeddings $\mathbf{z}_l = f_v(\mathbf{P}, \mathbf{x}_l)$.
%
This allows both $\mathbf{z}_l$ and $\mathbf{z}_h$ to be projected into a shared latent space, enabling cross-scale contrastive alignment.

To reduce the semantic gap between different scales, we employ two contrastive objectives: an instance loss $\mathcal{L}_{\text{inst}}$ and a prototype loss $\mathcal{L}_{\text{proto}}$.
The instance loss encourages alignment between each LR patch and its corresponding child HR patches.
Within a batch, HR patches that spatially fall within the region covered by an LR patch are treated as positives and others as negatives.
We define the instance-level loss as:
\begin{equation}
\mathcal{L}_{\mathrm{inst}}
=
-\frac{1}{|\mathcal{N}^{+}|}
\sum_{\mathbf{z}_h^{+} \in \mathcal{N}^{+}}
\log
\frac{
\exp\left( \mathbf{z}_l \cdot \mathbf{z}_h^{+} / \tau \right)
}{
\sum_{\mathbf{z}_h \in \mathcal{N}}
\exp\left( \mathbf{z}_l \cdot \mathbf{z}_h / \tau \right)
},
\end{equation}
where $\tau$ is the temperature parameter and $\mathbf{z}_h^{+}$ denotes a positive HR embedding.
$\mathcal{N}^{+}$ and $\mathcal{N}$ denote the positive and full patch sets within the batch, respectively.
However, if contrastive learning relies solely on the child (positive) vs. non-child (negative) relationship, negative patches that share the same pseudo-label as positive ones may be undesirably pushed apart.
%
In other words, the LR embedding may fail to capture the general semantic characteristics shared by positive HR patches.

To mitigate this, we introduce a prototype loss $\mathcal{L}_{\text{proto}}$, which encourages pseudo-label consistency across scales.
HR patches sharing the same pseudo-label are grouped into clusters $\{\mathcal{C}_k\}_{k=1}^{K}$, and the mean embedding of each cluster is used as its prototype:
\begin{equation}
\mathbf{p}_k = \frac{1}{|\mathcal{C}_k|} \sum_{\mathbf{z}_h \in \mathcal{C}_k} \mathbf{z}_h.
\end{equation}
Clusters containing at least one child (positive) patch are regarded as positive. 
Then, the prototype-level loss is defined as:
\begin{equation}
\mathcal{L}_{\mathrm{proto}}
=
-\frac{1}{|\mathcal{C}^{+}|}
\sum_{\mathbf{p}^{+} \in \mathcal{C}^{+}}
\log
\frac{
\exp\left( \mathbf{z}_l \cdot \mathbf{p}^{+} / \tau \right)
}{
\sum_{\mathbf{p} \in \mathcal{C}}
\exp\left( \mathbf{z}_l \cdot \mathbf{p} / \tau \right)
},
\end{equation}
where $\mathbf{p}^{+}$ and $\mathcal{C}^{+}$ denote the positive prototype and prototype sets within the batch, respectively.

The overall training objective of CS-VPT is the weighted combination of the two losses:
\begin{equation}
\mathcal{L}_{\text{CS-VPT}} = \lambda_{\text{inst}} \mathcal{L}_{\text{inst}} + \lambda_{\text{proto}} \mathcal{L}_{\text{proto}},
\end{equation}
where $\lambda_{\text{inst}}$ and $\lambda_{\text{proto}}$ balance the contributions of each term.
Together, these two losses enable robust cross-scale representation learning, effectively transferring diagnostic knowledge from HR to LR patches.
%
During training, only $\mathbf{P}$ and the projection layer of $f_v(\cdot)$ are optimized, while all other model parameters are frozen.
Section~\ref{subsec:effect_component} provides an ablation study examining the contribution of each loss.

\subsection{Coarse-to-Fine Informative Patch Selection
\\using VLM and LLM}
\label{sec:patch_selection}
\paragraph{Task-Relevant Concept Construction}

The $K$ pseudo-label concepts introduced for CS-VPT cover diverse
morphological patterns that may appear in each dataset. However, only
a subset of these concepts is important for a given downstream task. We therefore prompt an LLM to select
the top-$r$ task-relevant features in descending order of importance:
$\mathcal{F}_{r}
=
\left(f_1,f_2,\ldots,f_r\right)$,
where $f_1$ denotes the most important feature. Specifically, the LLM
is queried with:
\textit{``Among the $K{=}16$ visual features, list the top $r$ features that
are most important for \texttt{\{downstream task\}},
in descending order of importance.''}

For each selected feature, we construct multiple short phrases
describing its visual concept, following the text ensemble used for
HR pseudo-labeling. Their text embeddings are aggregated into a
feature-level representation $\mathbf{t}_j$ for feature $f_j$. The
resulting task-relevant text pool is defined as
$\mathcal{T}
=
\left\{
\mathbf{t}_1,\mathbf{t}_2,\ldots,\mathbf{t}_r
\right\}$,
where the features are ordered according to their importance to the
downstream task.
\begin{table*}[h]
\centering
\begingroup
\setlength{\tabcolsep}{1mm}
\small
\begin{tabular}{@{}l*{20}{c}@{}}
\toprule
\multirow{2}{*}{Method}
& \multicolumn{3}{c}{TCGA-BRCA}
& \multicolumn{3}{c}{TCGA-NSCLC}
& \multicolumn{3}{c}{TCGA-RCC}
& \multicolumn{3}{c}{BRCA\textsuperscript{*}}
& \multicolumn{3}{c}{UBC-OCEAN}
& \multicolumn{3}{c}{BRACS}
& \multicolumn{2}{c}{Avg.} \\
\cmidrule(lr){2-4}
\cmidrule(lr){5-7}
\cmidrule(lr){8-10}
\cmidrule(lr){11-13}
\cmidrule(lr){14-16}
\cmidrule(lr){17-19}
\cmidrule(l){20-21}
& Acc & AUC & $R$
& Acc & AUC & $R$
& Acc & AUC & $R$
& Acc & AUC & $R$
& Acc & AUC & $R$
& Acc & AUC & $R$
& Acc & AUC \\
\midrule
20$\times$ all
& 89.6 & \underline{93.1} & 100
& 91.7 & \underline{97.0} & 100
& 91.9 & 98.4 & 100
& 66.0 & \textbf{83.5} & 100
& \textbf{80.2} & \underline{94.2} & 100
& \underline{78.3} & 88.8 & 100
& \underline{83.0} & \underline{92.5} \\

20$\times$ 10\%
& 84.6 & 92.8 & 10
& 86.6 & 96.2 & 10
& 90.9 & 97.6 & 10
& 64.1 & 80.5 & 10
& 74.9 & 91.2 & 10
& 77.7 & 88.0 & 10
& 79.8 & 91.1 \\
\midrule
HDMIL
& 89.6 & \underline{93.1} & 10
& 91.2 & 95.8 & 10
& 92.5 & 98.0 & 10
& 62.8 & 80.4 & 10
& 77.3 & 92.7 & 10
& 77.4 & 89.1 & 10
& 81.8 & 91.5 \\

FOCUS
& \underline{89.7} & 92.7 & 10
& 91.2 & 95.8 & 10
& \underline{92.9} & 98.0 & 10
& 64.9 & 80.9 & 10
& 76.9 & 93.1 & 10
& 78.0 & \underline{89.9} & 10
& 82.3 & 91.7 \\

EmmPD
& 89.1 & 91.9 & 10
& 91.6 & 96.2 & 10
& 91.9 & 97.8 & 10
& 65.6 & \underline{83.1} & 10
& 76.2 & 92.8 & 10
& 74.8 & 86.3 & 10
& 81.5 & 91.4 \\

ZoomMIL
& 88.3 & 90.2 & 10
& \underline{91.8} & 96.5 & 10
& 90.9 & 97.8 & 10
& \underline{65.8} & 82.9 & 10
& 79.4 & 93.8 & 10
& 75.9 & 87.4 & 10
& 82.0 & 91.4 \\

HiVE-MIL
& 89.2 & 93.0 & 10
& 88.8 & 95.1 & 10
& 91.0 & \underline{98.5} & 10
& 62.3 & 78.8 & 10
& 76.5 & 93.0 & 10
& 74.8 & 87.1 & 10
& 80.4 & 90.9 \\

PATHS
& 89.3 & 91.7 & 10
& 90.6 & 96.0 & 10
& 92.5 & \underline{98.5} & 10
& 63.9 & 79.9 & 10
& 74.4 & 90.9 & 10
& 74.6 & 85.2 & 10
& 80.9 & 90.4 \\
\midrule
\textbf{Ours}
& \textbf{90.9} & \textbf{93.8} & 10
& \textbf{92.1} & \textbf{97.2} & 10
& \textbf{93.3} & \textbf{98.8} & 10
& \textbf{67.4} & 81.6 & 10
& \underline{79.7} & \textbf{94.7} & 10
& \textbf{81.0} & \textbf{91.1} & 10
& \textbf{84.1} & \textbf{92.9} \\
\bottomrule
\end{tabular}
\caption{Comparison of LanGuSTE with other methods on six classification tasks under the same budget of $R=10\%$. BRCA\textsuperscript{*} denotes molecular subtype classification on TCGA-BRCA. Avg. denotes the mean Acc and AUC across all tasks. The best and second-best results are shown in \textbf{bold} and \underline{underlined}, respectively.}
\label{tab:table_main}
\endgroup
\end{table*}
\begin{table*}[h]
\centering
\begingroup
\setlength{\tabcolsep}{1mm}
\small
\begin{tabular}{@{}l*{21}{c}@{}}
\toprule
\multirow{2}{*}{Method}
& \multicolumn{3}{c}{TCGA-BRCA}
& \multicolumn{3}{c}{TCGA-NSCLC}
& \multicolumn{3}{c}{TCGA-RCC}
& \multicolumn{3}{c}{BRCA\textsuperscript{*}}
& \multicolumn{3}{c}{UBC-OCEAN}
& \multicolumn{3}{c}{BRACS}
& \multicolumn{3}{c}{Avg.} \\
\cmidrule(lr){2-4}
\cmidrule(lr){5-7}
\cmidrule(lr){8-10}
\cmidrule(lr){11-13}
\cmidrule(lr){14-16}
\cmidrule(lr){17-19}
\cmidrule(l){20-22}
& Acc & AUC & $R$
& Acc & AUC & $R$
& Acc & AUC & $R$
& Acc & AUC & $R$
& Acc & AUC & $R$
& Acc & AUC & $R$
& Acc & AUC & $R$ \\
\midrule
20$\times$ all
& 89.6 & \underline{93.1} & 100
& 91.7 & \underline{97.0} & 100
& 91.9 & 98.4 & 100
& 66.0 & \textbf{83.5} & 100
& \textbf{80.2} & 94.2 & 100
& 78.3 & \underline{88.8} & 100
& 83.0 & \underline{92.5} & 100 \\

20$\times$ 10\%
& 84.6 & 92.8 & 10
& 86.6 & 96.2 & 10
& 90.9 & 97.6 & 10
& 64.1 & 80.5 & 10
& 74.9 & 91.2 & 10
& 77.7 & 88.0 & 10
& 79.8 & 91.1 & 10 \\
\midrule
HDMIL
& \underline{89.9} & \underline{93.1} & 70
& 91.2 & \underline{97.0} & 70
& 92.8 & \underline{98.8} & 60
& 65.5 & 81.7 & 70
& 78.5 & \underline{94.4} & 70
& 77.6 & 88.5 & 70
& 82.6 & 92.3 & 68 \\

FOCUS
& 89.4 & 92.9 & 15
& 92.0 & \underline{97.0} & 15
& 93.2 & \textbf{99.0} & 15
& 65.1 & 81.6 & 11
& 75.4 & 93.2 & 20
& 77.3 & 88.3 & 12
& 82.1 & 92.0 & 15 \\

EmmPD
& 89.4 & 93.0 & 35
& 91.2 & 96.2 & 35
& \underline{93.8} & 98.1 & 35
& \underline{66.7} & 82.3 & 36
& 77.9 & 94.2 & 32
& 76.6 & 85.7 & 34
& 82.6 & 91.6 & 35 \\

ZoomMIL
& 87.8 & 90.8 & 7
& \textbf{92.3} & 96.7 & 7
& 93.6 & 98.2 & 7
& 65.3 & 80.3 & 7
& 75.2 & 91.6 & 5
& 75.2 & 86.1 & 9
& 81.6 & 90.6 & 7 \\

HiVE-MIL
& \underline{89.9} & 92.4 & 60
& 91.4 & \textbf{97.2} & 60
& \textbf{94.4} & 98.6 & 60
& \underline{66.7} & \underline{83.4} & 60
& 78.7 & \underline{94.4} & 60
& \underline{78.5} & \underline{88.8} & 60
& \underline{83.3} & \underline{92.5} & 60 \\

PATHS
& 88.2 & 90.7 & 8
& 90.1 & 95.0 & 8
& 92.7 & 98.2 & 8
& 64.0 & 81.7 & 8
& 72.7 & 90.7 & 8
& 71.7 & 81.5 & 5
& 79.9 & 89.6 & 8 \\
\midrule
\textbf{Ours}
& \textbf{90.9} & \textbf{93.8} & 10
& \underline{92.1} & \textbf{97.2} & 10
& 93.3 & \underline{98.8} & 10
& \textbf{67.4} & 81.6 & 10
& \underline{79.7} & \textbf{94.7} & 10
& \textbf{81.0} & \textbf{91.1} & 10
& \textbf{84.1} & \textbf{92.9} & 10 \\
\bottomrule
\end{tabular}
\caption{Comparison of LanGuSTE with other methods on six classification tasks using each method's original patch-retention ratio. BRCA\textsuperscript{*} denotes molecular subtype classification on TCGA-BRCA. Avg. denotes the mean Acc, AUC, and $R$ across the six tasks. The best and second-best results are shown in \textbf{bold} and \underline{underlined}, respectively.}
\label{tab:table_original_ratio}
\endgroup
\end{table*}
\paragraph{Coarse-to-Fine Strategy}
To avoid processing all HR patches, we adopt a coarse-to-fine strategy that first identifies informative regions at low magnification and then refines the selection at high magnification.
LR patches are encoded using the VLM vision encoder $f_v(\cdot)$ together with the visual prompt $\mathbf{P}$ trained in the CS-VPT module, producing embeddings that implicitly retain HR information.
For each text representation $\mathbf{t}_j$ in the text pool $\mathcal{T}$, 
we compute its cosine similarities with the LR patch embeddings to construct a set of similarity maps:
$\mathbf{S} = \{ S_{1}, \dots, S_{r} \}$.
These similarity maps are passed to the patch selection module, which selects the top-$K_{1}\%$ LR patches.
For each selected LR patch, all corresponding HR patches are retrieved and encoded using the same vision encoder $f_v(\cdot)$.
Cosine similarities with respect to $\mathcal{T}$ are computed again, and the patch selection module is applied once more.
The top-$K_{2}\%$ HR patches per WSI are then selected as the final patch set.
\paragraph{Rank-Aware Patch Selection Module}
Given the similarity maps
$\mathbf{S}=\{S_{1},\ldots,S_{r}\}$, we incorporate the
LLM-derived importance ranking of the selected features into patch
scoring.
To progressively reflect the relative importance of the selected
features, we weight the similarity map of the $j$-th ranked feature as
$\widetilde{S}_{j}=\alpha^{j-1}S_{j}$, where
$\alpha\in(0,1)$ is a concept-rank decay weight.

To effectively integrate similarity information
across the selected features, we consider two complementary criteria:
\begin{enumerate}[label=(\roman*)]
\item Patches consistently scoring high across the maps capture general relevance to the task-related concepts.
\item Patches exhibiting high variation across the similarity maps show selective responses to particular key features and contain discriminative evidence.
\end{enumerate}
Accordingly, for each patch, we summarize the weighted similarities using the mean and standard deviation:
\begin{equation}
\mu
=
\operatorname{Mean}_{j=1}^{r}
\left(\widetilde{S}_{j}\right),
\qquad
\sigma
=
\operatorname{Std}_{j=1}^{r}
\left(\widetilde{S}_{j}\right).
\end{equation}
We standardize $\mu$ and $\sigma$ across patches and combine them into an informativeness score
$q
=
\mu+\beta\, \sigma,$
where $\beta$ controls the contribution of the variation term.
Finally, patches are ranked according to $q$, and the top-$K\%$ patches are retained.
The same module is applied with $K=K_1$ and $K=K_2$ for LR and HR selection, respectively.
\section{Experiments}
\subsection{Experimental Setup}
\subsubsection{Datasets}
We evaluate our method on multiple classification and survival prediction tasks using public datasets.
For histologic subtype classification, we use five cohorts: TCGA-BRCA (IDC vs. ILC), TCGA-NSCLC (LUAD vs. LUSC), TCGA-RCC (CCRCC vs. PRCC vs. CHRCC)~\cite{tcgareport}, UBC-OCEAN (CC vs. EC vs. HGSC vs. LGSC vs. MC)~\cite{ocean}, and BRACS (benign vs. atypical vs. malignant tumors)~\cite{BRACs}.
Additionally, we conduct molecular subtype classification on TCGA-BRCA (LumA vs. LumB vs. Basal vs. HER2).
For survival prediction, we use five TCGA datasets: BRCA, KIRC, KIRP, LUAD, and LUSC.

\subsubsection{Metrics}
For classification, we report accuracy (Acc) and area under the ROC curve (AUC). For survival prediction, we use the concordance index (C-index).
To evaluate the cross-scale alignment induced by CS-VPT, we report mean average precision at \textit{k} (mAP@\textit{k}) for $\textit{k}\in\{1,10\}$ in the retrieval analysis.
We additionally report the total wall-clock time, including patch cropping, feature extraction, model training, and inference, to assess the computational efficiency.

\subsubsection{Implementation Details}
We compare LanGuSTE with state-of-the-art patch-selection methods to evaluate the \emph
{quality of selection}. 
For a fair comparison, all methods share the CONCH~\cite{conch} vision encoder as the feature extractor, are evaluated with the same ABMIL~\cite{abmil} under 5-fold cross-validation, and retain the same patch ratios (Table~\ref{tab:table_main}).
In our pipeline, we use CONCH and GPT-5.5-Thinking~\cite{gpt5} as the VLM and LLM, respectively.
We set $\alpha=0.9$ for classification and $\alpha=0.7$ for survival prediction, with $r=8$, $K_1=30$, $K_2=10$, and $\beta=0.5$.
Further details are provided in the supplementary material.
\begin{figure*}[h]
\centering
\includegraphics[width=1\linewidth]{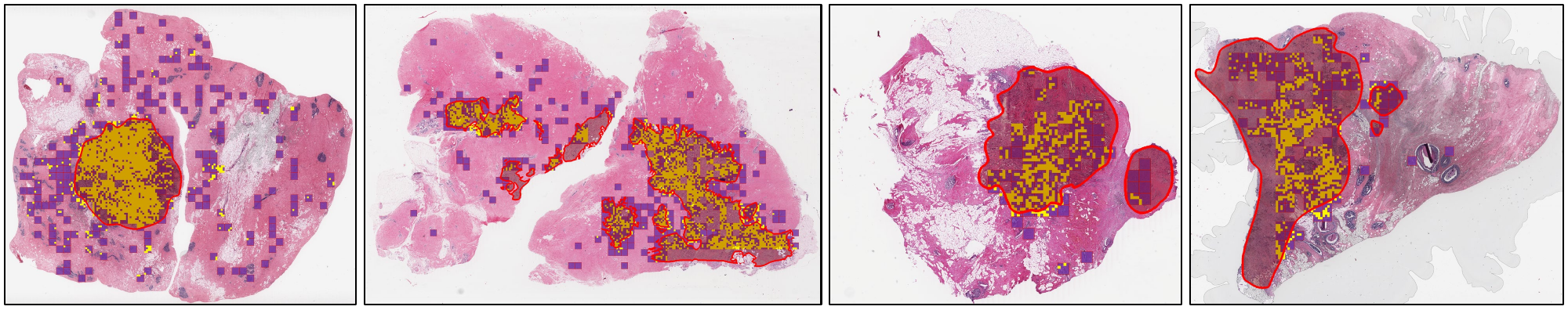}
\caption{Coarse-to-fine patch selection results by LanGuSTE on the TCGA-BRCA dataset. Purple boxes indicate candidate regions selected at the coarse-level stage (5$\times$); yellow boxes indicate patches selected at the fine-level stage (20$\times$) within those candidates. The red contours delineate the invasive carcinoma extent annotated by an expert pathologist. 
}
\label{fig:tissue_overlay}
\end{figure*}
\begin{table}[t]
\centering

\begingroup
\small
\setlength{\tabcolsep}{1mm}
\begin{tabular}{@{}lcccccc@{}}
\toprule
Method & BRCA & KIRC & KIRP & LUAD & LUSC & Avg. \\
\midrule
20$\times$ all  & 0.557 & 0.656 & 0.701 & 0.538 & 0.514 & 0.593 \\
20$\times$ 10\% & 0.547 & 0.676 & 0.695 & 0.530 & 0.500 & 0.590 \\
\midrule
HDMIL         & 0.589 & 0.666 & 0.627 & 0.538 & 0.512 & 0.586 \\
FOCUS         & 0.576 & 0.656 & 0.562 & 0.520 & 0.493 & 0.561 \\
EmmPD         & 0.579 & \underline{0.693} & 0.663 & \textbf{0.559} & \underline{0.544} & \underline{0.608} \\
ZoomMIL       & 0.577 & 0.628 & \underline{0.708} & \underline{0.551} & 0.522 & 0.597 \\
HiVE-MIL      & 0.566 & 0.685 & 0.668 & \underline{0.551} & \textbf{0.571} & \underline{0.608} \\
PATHS         & \underline{0.612} & 0.659 & 0.620 & 0.533 & 0.543 & 0.593 \\
\midrule
\textbf{Ours} & \textbf{0.614} & \textbf{0.695} & \textbf{0.714} & 0.550 & 0.540 & \textbf{0.623} \\
\bottomrule
\end{tabular}
\caption{Comparison of survival prediction performance across five TCGA cohorts.
Avg. denotes the mean across the five cohorts.
The best and second-best results are shown in \textbf{bold} and
\underline{underlined}, respectively.}
\label{tab:survival_prediction}
\endgroup
\end{table}
%
\subsection{Quantitative Results}
We compare LanGuSTE with two reference settings and six patch-selection methods across six classification tasks.
``20$\times$ all'' uses all HR patches, whereas ``20$\times$ 10\%'' uses 10\% of randomly sampled HR patches.
All patch-selection methods use the same training split as CS-VPT. After training, each selection model is frozen and applied to the full dataset, and the resulting patch sets are evaluated using ABMIL under the same cross-validation protocol. We denote the retained patch ratio by $R$. \\
Table~\ref{tab:table_main} compares all patch-selection methods under the same budget of $R=10\%$. LanGuSTE achieves the highest average accuracy and AUC of 84.1\% and 92.9\%, respectively. It obtains the best AUC on five of the six tasks.
Notably, LanGuSTE outperforms exhaustive 20$\times$ processing in average metrics while using only one-tenth of the HR patches.
Table~\ref{tab:table_original_ratio} further evaluates each method using its originally proposed patch-retention ratio.
LanGuSTE achieves the best average performance while using only 10\% of the HR patches, compared with 60\% for the second-best HiVE-MIL and 100\% for exhaustive 20$\times$ processing.
These results demonstrate that LanGuSTE provides a strong balance between classification performance and HR patch-processing efficiency.

\begin{figure}[h]
\centering
\includegraphics[width=0.95\linewidth]{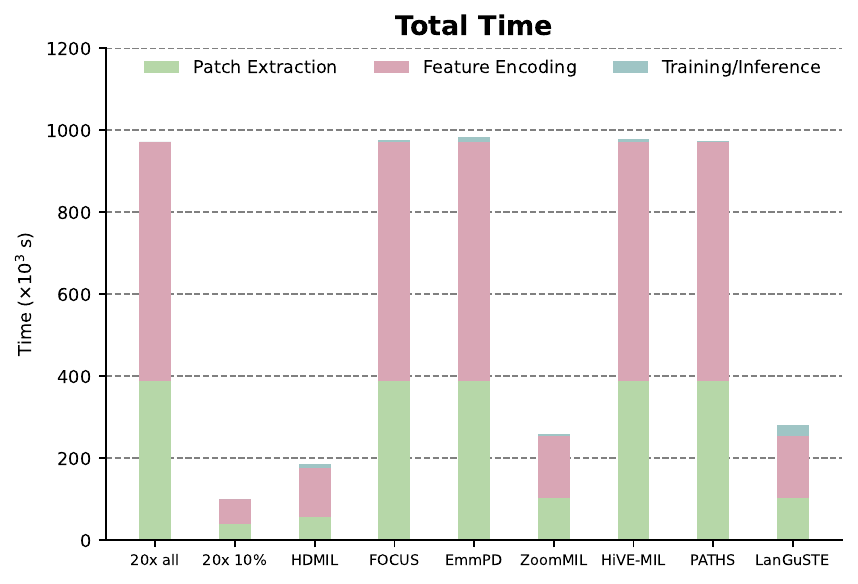}
\caption{Breakdown of WSI processing time across five datasets. Note that patch extraction and per-patch feature encoding are the major bottlenecks, and LanGuSTE significantly reduces them with low overhead.}
\label{timeplot}
\end{figure}

\subsubsection{Survival Prediction}
Table~\ref{tab:survival_prediction} reports survival prediction performance across five TCGA cohorts.
LanGuSTE achieves the highest average C-index of 0.623, outperforming both exhaustive 20$\times$
processing and the existing patch-selection methods.
These results indicate that LanGuSTE preserves prognostically relevant information and generalizes beyond classification tasks.

\subsubsection{Runtime Analysis}
Figure~\ref{timeplot} illustrates the average processing time required by each method across five classification datasets with an NVIDIA RTX 4090 GPU. 
Although preprocessing all HR patches requires approximately 11 days in our setting, LanGuSTE limits HR patch processing to the top-$K_1=30\%$ of LR regions,
bringing the total processing time down to less than four days, approximately a threefold improvement in efficiency.
As the size of dataset 
increases, this efficiency gain becomes even more critical. 
Although our framework introduces additional steps such as LR patch preprocessing, cross-scale alignment, and similarity matrix computation, these components incur substantially less overhead compared to HR patch processing. Consequently, LanGuSTE maintains a clear advantage in terms of time cost.
\begin{table*}[h]
\centering

\begingroup
\small
\setlength{\tabcolsep}{1mm}
\renewcommand{\arraystretch}{1.15}
\begin{tabular}{@{}ccc*{7}{cc}@{}}
\toprule
\multirow{2}{*}{$\mathcal{L}_{\mathrm{inst}}$}
& \multirow{2}{*}{$\mathcal{L}_{\mathrm{proto}}$}
& \multirow{2}{*}{C2F Sel.}
& \multicolumn{2}{c}{TCGA-BRCA}
& \multicolumn{2}{c}{TCGA-NSCLC}
& \multicolumn{2}{c}{TCGA-RCC}
& \multicolumn{2}{c}{BRCA\textsuperscript{*}}
& \multicolumn{2}{c}{UBC-OCEAN}
& \multicolumn{2}{c}{BRACS}
& \multicolumn{2}{c}{Avg.} \\
\cmidrule(lr){4-5}
\cmidrule(lr){6-7}
\cmidrule(lr){8-9}
\cmidrule(lr){10-11}
\cmidrule(lr){12-13}
\cmidrule(l){14-15}
\cmidrule(l){16-17}
& &
& Acc & AUC
& Acc & AUC
& Acc & AUC
& Acc & AUC
& Acc & AUC
& Acc & AUC
& Acc & AUC \\
\midrule

\xmark & \xmark & \xmark
& 88.2 & 92.3
& 89.8 & 96.5
& \textbf{93.3} & 98.7
& 64.9 & \textbf{81.7}
& \textbf{80.8} & 94.0
& 77.0 & 87.8
& 82.3 & 91.8 \\

\checkmark & \xmark & \xmark
& 90.6 & 92.8
& 90.8 & 96.8
& 91.8 & 98.5
& 66.2 & 81.6
& 79.8   & 93.5
& 78.8 & 89.2
& 83.0 & 92.1 \\

\checkmark & \checkmark & \xmark
& 90.1 & 92.9
& 91.5 & 97.1
& 92.7 & 98.3
& 65.2 & 81.3
& 78.7 & 94.0
& 77.6 & 89.7
& 82.6 & 92.2 \\

\checkmark & \xmark & \checkmark
& 90.8 & 93.7
& 91.7 & 97.1
& 92.4 & 98.6
& 66.8 & 81.6
& 79.3 & 94.1
& \textbf{81.0} & 90.3
& 83.7 & 92.6 \\

\checkmark & \checkmark & \checkmark
& \textbf{90.9} & \textbf{93.8}
& \textbf{92.1} & \textbf{97.2}
& \textbf{93.3} & \textbf{98.8}
& \textbf{67.4} & 81.6
& 79.7 & \textbf{94.7}
& \textbf{81.0} & \textbf{91.1}
& \textbf{84.1} & \textbf{92.9} \\

\bottomrule
\end{tabular}
\caption{Ablation study of each component in LanGuSTE. BRCA\textsuperscript{*} denotes molecular subtype classification on TCGA-BRCA. Avg. denotes the mean Acc and AUC across the six tasks. The best result is marked in \textbf{bold}.}
\label{tab:table_ablation}
\endgroup
\end{table*}
%
%
%
%
\subsection{Qualitative Results}
Figure~\ref{fig:tissue_overlay} visualizes the patches selected by LanGuSTE overlaid with pathologist-provided annotation masks~\cite{hishi} for four representative WSIs.
The final HR patches are concentrated within or near the annotated diagnostic regions, showing strong agreement with expert annotations.
Although the coarse-level selection may initially include some irrelevant areas among the top-$K_1\%$ candidates, the fine-level refinement effectively removes such regions and retains patches containing the most relevant morphological evidence.
These results qualitatively demonstrate the effectiveness of the coarse-to-fine selection strategy.
%
\subsection{Ablation Study}
\label{subsec:effect_component}
Table~\ref{tab:table_ablation} evaluates the contributions of each component in our framework.
`C2F Sel.' denotes the full coarse-to-fine selection process, in which patches are refined at both the coarse- and fine-level. 
In contrast, in the variant without `C2F Sel.', we directly select the top-$K_2\%$ coarse-level patches, using all of their HR child patches without further refinement. 
Comparing the configurations with and without C2F Sel. (i.e., rows 2 vs.\ 4 and rows 3 vs.\ 5) shows that fine-level refinement consistently improves the average performance by filtering uninformative patches.
Adding $\mathcal{L}_{\mathrm{proto}}$ without C2F Sel. yields mixed effects (rows 2 vs.\ 3), whereas its inclusion in the complete model further improves the average AUC from 92.6\% to 92.9\% (rows 4 vs.\ 5).
Overall, the complete model achieves the best average accuracy and AUC,
demonstrating the complementary contributions of cross-scale alignment and the coarse-to-fine strategy.
Additionally, we conduct ablation studies on the hyperparameters $r$, $K_{1}$, $\alpha$, and $\beta$, as well as the choice of VLM. Results are provided in the supplementary material.
\begin{table}[t]
\centering

\begingroup
\small
\setlength{\tabcolsep}{1mm}
\renewcommand{\arraystretch}{1.08}

\begin{tabular}{@{}l*{4}{cc}@{}}
\toprule
\multirow{2}{*}{Query}
& \multicolumn{2}{c}{BRCA(@\textit{k})}
& \multicolumn{2}{c}{NSCLC(@\textit{k})}
& \multicolumn{2}{c}{RCC(@\textit{k})}
& \multicolumn{2}{c}{BRACS(@\textit{k})} \\
\cmidrule(lr){2-3}
\cmidrule(lr){4-5}
\cmidrule(lr){6-7}
\cmidrule(l){8-9}
& @1 & @10
& @1 & @10
& @1 & @10
& @1 & @10 \\
\midrule

5$\times$
& 35.0 & 28.0
& 54.3 & 40.9
& 62.0 & 53.3
& 67.5 & 59.5 \\

\makecell[l]{5$\times$ w/\\CS-VPT}
& \textbf{59.7} & \textbf{54.0}
& \textbf{62.4} & \textbf{51.2}
& \textbf{75.0} & \textbf{66.4}
& \textbf{76.6} & \textbf{69.3} \\

\bottomrule
\end{tabular}

\caption{Cross-scale retrieval results for evaluating
semantic alignment between 5$\times$ and 20$\times$ representations. Mean Average Precision values are reported as percentages.}
\label{tab:table_retrieval}
\endgroup
\end{table}
\subsection{Functional Validation and Reliability Analysis}
\subsubsection{Cross-Scale Semantic Alignment}
To examine whether CS-VPT transfers fine-grained information from 20$\times$ patches to 5$\times$ representations, we conduct cross-scale semantic retrieval. 
Each 5$\times$ patch is used as a query to retrieve 20$\times$ patches from the same WSI, and a retrieval is considered relevant when its pseudo-label matches that of any child patch.
As shown in Table~\ref{tab:table_retrieval}, CS-VPT consistently improves mAP@1 and mAP@10 across all datasets, supporting improved alignment between the two magnifications.
Qualitative examples of 5$\times$ patches selected using 20$\times$ textual descriptions are provided in the supplementary material.
\begin{figure}[h]
\centering
\includegraphics[width=0.95\linewidth]{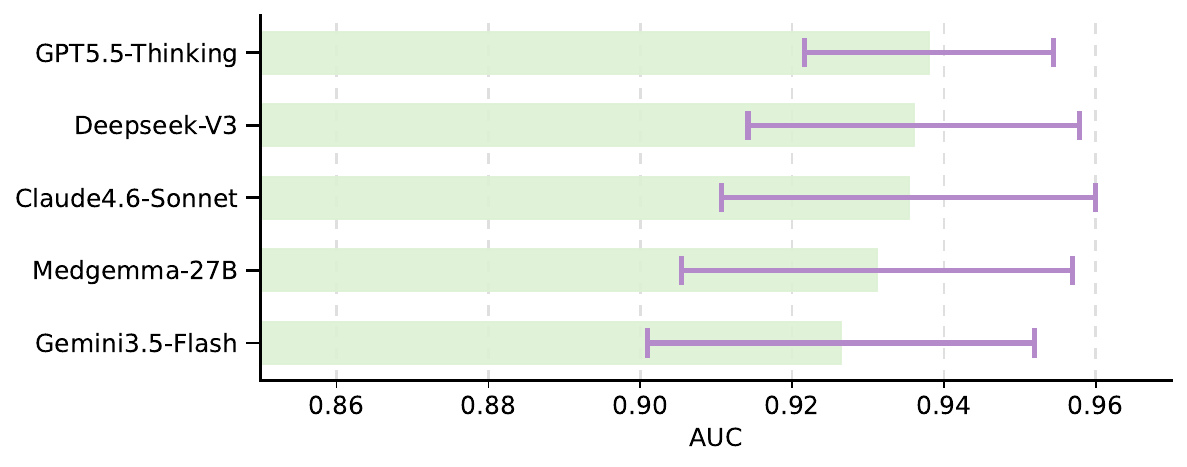}
\caption{Performance comparison of LanGuSTE using various LLMs. Error bars indicate variation
across cross-validation folds.}
\label{llm_comparison}
\end{figure}
\subsubsection{LLM Robustness and Clinical Plausibility}
Since LanGuSTE uses LLM-derived knowledge to construct task-relevant concepts, we assess its sensitivity to the choice of LLM.
We replace GPT5.5-Thinking with four alternative LLMs \cite{deepseek,anthropic2026sonnet46,medgemma,gemini} while keeping all other components fixed and evaluate on TCGA-BRCA histologic subtype classification.
As shown in
Figure~\ref{llm_comparison}, all LLMs yield consistently high AUC values, indicating robustness of the framework.
Furthermore, a pathologist independently reviewed the LLM-derived knowledge and judged the selected concepts and their task relevance to be pathologically plausible.
Textual resources are provided in the supplementary material.
\section{Conclusion}
In this paper, we introduce LanGuSTE for efficiently encoding large-scale WSIs, addressing a critical need in CPath. Leveraging the capabilities of pre-trained VLMs and LLMs, LanGuSTE adopts a coarse-to-fine patch selection strategy that identifies diagnostically relevant regions without the need to process all patches. LanGuSTE reduces processing time by approximately threefold on average across datasets, while achieving superior diagnostic performance compared to full-resolution and other state-of-the-art baselines by processing only a small subset of informative patches. Quantitative and qualitative evaluations highlight the effectiveness and scalability of LanGuSTE. 
Since multiple components of the framework depend on the quality of pre-trained VLMs and LLMs, exploring its performance with different models presents an interesting avenue for future work.

\bibliography{aaai2027}

\clearpage
\setcounter{section}{0}
\setcounter{subsection}{0}
\setcounter{subsubsection}{0}
\renewcommand{\thesection}{\arabic{section}}
\renewcommand{\thesubsection}{\thesection.\arabic{subsection}}
\renewcommand{\thesubsubsection}{\thesubsection.\arabic{subsubsection}}
\twocolumn[
\begin{center}
    {\huge\bfseries Supplementary Material}\par\vspace{0.5em}
    {\Large\bfseries LanGuSTE: Language-Guided Coarse-to-Fine Patch Selection \\for Efficient Whole Slide Image Analysis}\par
\end{center}
\vspace{1em}
]

\section{Dataset Statistics}

For our experiments, we used five publicly available datasets including TCGA-BRCA, TCGA-NSCLC, TCGA-RCC~\cite{tcgareport}, UBC-OCEAN~\cite{ocean}, and BRACS~\cite{BRACs}.
All WSIs were preprocessed by removing background regions using Otsu's thresholding, and image patches were extracted with a fixed size of 256×256 pixels.
We utilized 5× magnification for low-resolution patches and 20× magnification for high-resolution patches.
Table~\ref{tab:dataset_stats} summarizes the number of slides and extracted patches at low and high magnifications.

\begin{table}[h]
\centering
\setlength{\tabcolsep}{0.9mm}
\begin{tabular}{lccc}
\toprule
\textbf{Dataset} & \textbf{\# Slides} & \textbf{\# 5$\times$ patches} & \textbf{\# 20$\times$ patches} \\
\midrule
TCGA-BRCA  & 1,003 & 539,327 & 7,716,660 \\
TCGA-NSCLC & 3,129 & 732,882 & 11,701,195 \\
TCGA-RCC   & 3,066 & 1,019,700 & 16,354,162 \\
UBC-OCEAN  & 536 & 408,846 & 6,031,175 \\
BRACS      & 546 & 341,104 & 5,167,919 \\
\bottomrule
\end{tabular}
\caption{Number of slides and patches for each dataset.}
\label{tab:dataset_stats}
\end{table}



\section{Implementation Details}
\subsubsection*{Cross-Scale Visual Prompt Tuning (CS-VPT)} Table~\ref{tab:csvpt_details} summarizes the training details of CS-VPT used for preserving fine-grained diagnostic features at a coarse level.
\begin{table}[h]
\centering
\small
\setlength{\tabcolsep}{6pt}
\begin{tabular}{ll}
\toprule
Configuration & Value \\
\midrule
Training WSIs per class & 5 \\
Learnable prompts & 30 \\
Epochs & 200 \\
Loss weights ($\lambda_{inst}$, $\lambda_{proto}$) & 1.0, 0.5 \\
Learning rate & 1e-4 \\
Optimizer & Adam \\
Batch size & 64 \\
\bottomrule
\end{tabular}
\caption{Training details of CS-VPT for preserving fine-grained diagnostic features at a coarse level.}
\label{tab:csvpt_details}
\end{table}


\newpage
\subsubsection*{ABMIL Training}
Table~\ref{tab:abmil_classification} and ~\ref{tab:abmil_survival} summarize the training procedure of ABMIL for evaluating the quality of selection, which is shared across all datasets and patch-selection methods.
\begin{table}[h]
\centering
\small
\setlength{\tabcolsep}{6pt}
\begin{tabular}{ll}
\toprule
Configuration & Value \\
\midrule
Data split & 5-fold CV, stratified (8:1:1) \\
Epochs & 200 \\
Learning rate & 1e-4 \\
Loss function & Cross Entropy Loss\\
Optimizer & Adam \\
Early stopping & Patience 30 (val. loss) \\
Weight decay & 1e-4 \\
LR scheduler & CosineAnnealingLR \\
\bottomrule
\end{tabular}
\caption{Training details of ABMIL for the classification tasks.}
\label{tab:abmil_classification}
\end{table}
\begin{table}[h]
\centering
\small
\setlength{\tabcolsep}{6pt}
\begin{tabular}{ll}
\toprule
Configuration & Value \\
\midrule
Data split & 5-fold CV, stratified (8:1:1) \\
Epochs & 300 \\
Learning rate & 1e-4 \\
Batch size & 64 \\
Loss function & Cox Partial Likelihood Loss \\
Optimizer & Adam \\
Early stopping & Patience 30 (val. loss) \\
Weight decay & 1e-4 \\
LR scheduler & CosineAnnealingLR \\
\bottomrule
\end{tabular}
\caption{Training details of ABMIL for the survival prediction tasks.}
\label{tab:abmil_survival}
\end{table}


\onecolumn
\section{Further Experimental Analysis}
This section provides additional experiments and analyses that complement the results presented in the main paper. We first report survival prediction results using the original patch-retention ratio of each competing method. We then examine the sensitivity of LanGuSTE to its major hyperparameters, including the coarse-level retention ratio $K_1$, the number of task-relevant concepts $r$, the concept-rank decay factor $\alpha$, and the standard-deviation weight $\beta$. We additionally evaluate the effect of the pathology VLM backbone and provide qualitative analyses of cross-scale semantic retrieval and coarse-to-fine patch selection. Together, these results further assess the robustness, effectiveness, and interpretability of the proposed framework.


\subsection{Additional Survival Prediction Result}
Table~\ref{tab:survival_original} compares the survival prediction performance of each method~\cite{hdmil,focus,emmpd,zoommil,hivemil,paths} using its originally proposed patch-retention ratio.
LanGuSTE achieves the highest average C-index while retaining only 10\% of the HR patches.
In particular, it obtains the best performance on the BRCA, KIRC, and KIRP cohorts.
Compared with exhaustive 20$\times$ processing, LanGuSTE improves the average C-index from 0.593 to 0.623 while reducing the retained HR patches by 90\%. It also outperforms the strongest competing baseline, FOCUS, by 0.011 in average C-index while using a lower patch-retention ratio (10\% vs. 15\%). Although several baselines achieve higher results on individual cohorts, LanGuSTE provides the best overall performance and a favorable trade-off between prognostic accuracy and computational efficiency.
\begin{table}[h]
\centering

\begingroup
\small
\setlength{\tabcolsep}{1mm}
\begin{tabular}{@{}lccccccc@{}}
\toprule
Method & BRCA & KIRC & KIRP & LUAD & LUSC & Avg. & $R$ \\
\midrule
20$\times$ all  & 0.557 & 0.656 & \underline{0.701} & 0.538 & 0.514 & 0.593 & 100 \\
20$\times$ 10\% & 0.547 & 0.676 & 0.695 & 0.530 & 0.500 & 0.590 & 10 \\
\midrule
HDMIL         & 0.574 & 0.643 & 0.591 & \textbf{0.575} & 0.517 & 0.580 & 66 \\
FOCUS         & \underline{0.594} & 0.667 & 0.706 & 0.548 & 0.546 & \underline{0.612} & 15 \\
EmmPD         & 0.537 & \underline{0.685} & 0.588 & \underline{0.568} & 0.507 & 0.577 & 35 \\
ZoomMIL       & 0.581 & 0.643 & 0.577 & 0.540 & 0.547 & 0.578 & 7 \\
HiVE-MIL      & 0.564 & 0.651 & 0.664 & 0.561 & \underline{0.553} & 0.599 & 67 \\
PATHS         & 0.576 & 0.630 & 0.577 & 0.531 & \textbf{0.555} & 0.574 & 15 \\
\midrule
\textbf{Ours} & \textbf{0.614} & \textbf{0.695} & \textbf{0.714} & 0.550 & 0.540 & \textbf{0.623} & 10 \\
\bottomrule
\end{tabular}
\caption{Comparison for survival prediction using each method's original patch-retention ratio across five TCGA cohorts.
Metrics are C-index under 5-fold cross-validation, and Avg.\ denotes the mean across all cohorts.
$R$ denotes the average of the retained patch ratio.
The best and second-best results are shown in \textbf{bold} and \underline{underlined}, respectively.
}
\label{tab:survival_original}
\endgroup
\end{table}

\newcommand{\survnote}{Metrics are C-index under 5-fold cross-validation. Avg.\ denotes the mean across all cohorts. The best and second-best results are shown in \textbf{bold} and \underline{underlined}, respectively.}

\clearpage
\onecolumn
\subsection{Ablation Study on the Coarse-Level Retention Ratio $K_1$}

$K_1$ determines the proportion of 5$\times$ patches retained at the coarse stage.
All 20$\times$ child patches corresponding to these selected regions are subsequently encoded and passed to the fine-level selection stage.
A small $K_1$ improves efficiency but may discard diagnostically relevant regions before HR analysis, whereas a large $K_1$ increases computation and may introduce uninformative candidates.
We sweep $K_1\in\{10,20,30,40,50\}$ while fixing the fine-level retention ratio to $K_2=10$, with $r=8$, $\beta=0.5$, and $\alpha$ set to its task-family-specific value ($0.9$ for classification and $0.7$ for survival prediction).

\subsubsection{Classification}
Classification performance generally improves as $K_1$ increases from 10\% to 30--40\%, indicating that an overly restrictive coarse-stage selection can prematurely remove useful regions. Increasing $K_1$ from 10 to 30 improves the average accuracy from 82.6\% to 84.1\% and the average AUC from 92.2\% to 92.9\%. The best average accuracy is obtained at $K_1=40$ with 84.2\%, while $K_1=30$ is only 0.1 points lower and achieves the same best average AUC of 92.9\%. Increasing the ratio further to 50\% does not provide additional benefit, reducing the average AUC to 92.6\%. Although the optimal ratio varies across individual cohorts, the results show that retaining approximately 30--40\% of the coarse regions provides sufficient morphological coverage, whereas retaining either too few or unnecessarily many regions is less effective. We adopt $K_1=30$ because it achieves nearly the best aggregate classification performance with a lower HR-processing cost than $K_1=40$.

\begin{table}[!htb]
\centering
\begingroup
\setlength{\tabcolsep}{2.2mm}
\begin{tabular}{@{}c*{14}{c}@{}}
\toprule
\multirow{2}{*}{$K_1$}
& \multicolumn{2}{c}{TCGA-BRCA}
& \multicolumn{2}{c}{TCGA-NSCLC}
& \multicolumn{2}{c}{TCGA-RCC}
& \multicolumn{2}{c}{BRCA\textsuperscript{*}}
& \multicolumn{2}{c}{UBC-OCEAN}
& \multicolumn{2}{c}{BRACS}
& \multicolumn{2}{c}{Avg.} \\
\cmidrule(lr){2-3}
\cmidrule(lr){4-5}
\cmidrule(lr){6-7}
\cmidrule(lr){8-9}
\cmidrule(lr){10-11}
\cmidrule(lr){12-13}
\cmidrule(l){14-15}
& Acc & AUC
& Acc & AUC
& Acc & AUC
& Acc & AUC
& Acc & AUC
& Acc & AUC
& Acc & AUC \\
\midrule
10
& 90.1 & 92.9
& 91.5 & \underline{97.1}
& 92.7 & 98.3
& 65.2 & 81.3
& 78.7 & 94.0
& 77.6 & 89.7
& 82.6 & 92.2 \\

20
& \textbf{91.0} & 93.4
& 91.7 & \underline{97.1}
& 92.7 & 98.7
& 66.1 & \textbf{82.2}
& \textbf{80.2} & 94.2
& 79.9 & 90.7
& 83.6 & \underline{92.7} \\

\rowcolor{blue!8} 30
& \underline{90.9} & \textbf{93.8}
& \textbf{92.1} & \textbf{97.2}
& \underline{93.3} & \underline{98.8}
& 67.4 & \underline{81.6}
& 79.7 & \textbf{94.7}
& 81.0 & \textbf{91.1}
& \underline{84.1} & \textbf{92.9} \\

40
& 90.6 & \textbf{93.8}
& \underline{91.8} & \textbf{97.2}
& \textbf{94.2} & \textbf{98.9}
& \underline{67.7} & 82.0
& 79.3 & 94.4
& \underline{81.7} & \underline{90.9}
& \textbf{84.2} & \textbf{92.9} \\

50
& \textbf{91.0} & \underline{93.7}
& 90.7 & 97.0
& 92.2 & 98.3
& \textbf{67.9} & \underline{81.6}
& \underline{79.8} & \underline{94.5}
& \textbf{82.1} & 90.5
& 84.0 & 92.6 \\
\bottomrule
\end{tabular}
\caption{Effect of coarse-level ratio $K_1$ on six classification tasks.
BRCA\textsuperscript{*} denotes molecular subtype classification on TCGA-BRCA.
Metrics are Acc\,/\,AUC (\%) under 5-fold cross-validation, and Avg.\ denotes the mean Acc and AUC across all tasks.
The best and second-best results are shown in \textbf{bold} and \underline{underlined}, respectively, and the highlighted row indicates the default setting used in the main paper.
}
\label{tab:k1_ablation}
\endgroup
\end{table}

\subsubsection{Survival Prediction}
The effect of $K_1$ is more pronounced for survival prediction. $K_1=30$ achieves the highest average C-index of 0.623, outperforming the second-best setting, $K_1=20$, by 0.019. It also yields the best performance on BRCA, KIRP, and LUAD and the second-best result on LUSC. Performance decreases at both ends of the sweep, reaching 0.595 at $K_1=10$ and 0.579 at $K_1=50$. This non-monotonic trend suggests that a small coarse-level ratio may omit spatially sparse prognostic evidence, while an excessively large ratio introduces less informative regions that weaken the subsequent fine-level refinement. Overall, $K_1=30$ provides the most favorable balance between prognostic performance and processing efficiency and is therefore used in all experiments.

\begin{table}[!htb]
\centering
\begingroup
\setlength{\tabcolsep}{2.2mm}
\begin{tabular}{@{}ccccccc@{}}
\toprule
$K_1$ & BRCA & KIRC & KIRP & LUAD & LUSC & Avg. \\
\midrule
10
& 0.539
& \underline{0.701}
& 0.643
& 0.510
& \textbf{0.581}
& 0.595 \\

20
& 0.568
& 0.685
& \underline{0.706}
& \underline{0.536}
& 0.526
& \underline{0.604} \\

\rowcolor{blue!8} 30
& \textbf{0.614}
& 0.695
& \textbf{0.714}
& \textbf{0.550}
& \underline{0.540}
& \textbf{0.623} \\

40
& \underline{0.586}
& 0.699
& 0.657
& 0.528
& 0.515
& 0.597 \\

50
& 0.555
& \textbf{0.709}
& 0.588
& 0.517
& 0.527
& 0.579 \\
\bottomrule
\end{tabular}
\caption{Effect of coarse-level ratio $K_1$ on five survival prediction tasks.
Metrics are C-index under 5-fold cross-validation, and Avg. denotes the mean across all cohorts.
The best and second-best results are shown in \textbf{bold} and \underline{underlined}, respectively, and the highlighted row indicates the default setting used in the main paper.
}
\label{tab:k1_survival_ablation}
\endgroup
\end{table}

\clearpage
\subsection{Ablation Study on the Number of Task-Relevant Concepts $r$}
$r$ determines how many of the $K=16$ pseudo-label concepts are retained as the task-relevant text pool $\mathcal{T}$, where $r=16$ keeps the full vocabulary and thus disables concept selection while retaining the LLM-derived ranking through $\alpha$. We sweep $r\in\{1,2,4,8,16\}$ with $K_1=30$, $\beta=0.5$ and $\alpha$ set to its per-task-family value.

\subsubsection{Classification}
Too few concepts are clearly harmful: $r\le2$ loses 2.6--3.6 points of average accuracy relative to $r=8$, with the gap reaching 9.5\% on UBC-OCEAN and 7.1\% on BRACS, where a small set of dominant concepts cannot separate subtypes that share coarse tissue architecture. Retaining the full vocabulary is also suboptimal, costing 0.9\% average accuracy and 2.9\% average AUC. The AUC drop is concentrated on BRACS (91.1\% to 74.9\%), while UBC-OCEAN in fact improves slightly, indicating that task-irrelevant concepts act as strong distractors on some cohorts rather than degrading all tasks uniformly. $r=8$ attains the best average on both metrics.
\begin{table*}[!htb]
\centering
\setlength{\tabcolsep}{4.5pt}
\begin{tabular}{c cc cc cc cc cc cc cc}
\toprule
& \multicolumn{2}{c}{TCGA-BRCA} & \multicolumn{2}{c}{TCGA-NSCLC} & \multicolumn{2}{c}{TCGA-RCC} & \multicolumn{2}{c}{BRCA\textsuperscript{*}} & \multicolumn{2}{c}{UBC-OCEAN} & \multicolumn{2}{c}{BRACS} & \multicolumn{2}{c}{Avg.} \\
\cmidrule(lr){2-3}\cmidrule(lr){4-5}\cmidrule(lr){6-7}\cmidrule(lr){8-9}\cmidrule(lr){10-11}\cmidrule(lr){12-13}\cmidrule(lr){14-15}
$r$ & Acc & AUC & Acc & AUC & Acc & AUC & Acc & AUC & Acc & AUC & Acc & AUC & Acc & AUC \\
\midrule
1 & \underline{90.9} & \underline{93.8} & 90.6 & 96.3 & 92.1 & 98.6 & 65.2 & \underline{83.1} & 70.2 & 88.5 & 73.9 & 83.3 & 80.5 & 90.6 \\
2 & 90.4 & \textbf{94.0} & 91.4 & \textbf{97.3} & \underline{93.3} & 98.6 & \underline{66.6} & \textbf{83.8} & 72.9 & 92.2 & 74.1 & 84.2 & 81.5 & 91.7 \\
4 & \textbf{91.0} & \textbf{94.0} & \textbf{92.6} & 97.1 & \textbf{94.6} & 98.7 & \textbf{67.4} & 82.5 & 76.8 & 93.1 & 77.6 & \underline{86.9} & \underline{83.3} & \underline{92.1} \\
\rowcolor{blue!8} 8 & \underline{90.9} & \underline{93.8} & \underline{92.1} & \underline{97.2} & \underline{93.3} & \underline{98.8} & \textbf{67.4} & 81.6 & \underline{79.7} & \underline{94.7} & \textbf{81.0} & \textbf{91.1} & \textbf{84.1} & \textbf{92.9} \\
16 & 90.2 & \underline{93.8} & 90.9 & 96.9 & 92.1 & \textbf{98.9} & 64.7 & 80.5 & \textbf{82.2} & \textbf{94.8} & \underline{79.0} & 74.9 & 83.2 & 90.0 \\
\bottomrule
\end{tabular}
\caption{Effect of the number of task-relevant concepts $r$ selected from the $K=16$ pseudo-label concepts.
BRCA\textsuperscript{*} denotes molecular subtype classification on TCGA-BRCA.
Metrics are Acc\,/\,AUC (\%) under 5-fold cross-validation, and Avg.\ denotes the mean Acc and AUC across all tasks.
The best and second-best results are shown in \textbf{bold} and \underline{underlined}, respectively, and the highlighted row indicates the default setting used in the main paper.
}
\label{tab:r_ablation_subtype}
\end{table*}

\subsubsection{Survival Prediction}
The trend differs from classification: $r=1$ attains the best average C-index (0.627), marginally ahead of $r=8$ (0.623), and performance varies non-monotonically over a narrow range (0.596--0.627). No setting is consistently best across cohorts---$r=1$ leads on LUSC, $r=8$ on BRCA and LUAD, and $r=16$ on KIRC---suggesting that the differences reflect cohort-level noise rather than a systematic effect of $r$. This is consistent with prognostic signal being carried by a few strongly discriminative concepts rather than by broad morphological coverage, mirroring the smaller $\alpha$ preferred for survival. Since $r=8$ is within 0.4 points of the best setting and is required for classification, we use $r=8$ for both task families.
\begin{table}[!htb]
\centering

\begingroup
\setlength{\tabcolsep}{2.2mm}
\begin{tabular}{@{}lcccccc@{}}
\toprule
$r$ & BRCA & KIRC & KIRP & LUAD & LUSC & Avg. \\
\midrule
1 & \underline{0.594} & \underline{0.706} & \textbf{0.714} & 0.526 & \textbf{0.594} & \textbf{0.627} \\
2 & 0.560 & 0.701 & 0.654 & 0.516 & \underline{0.547} & 0.596 \\
4 & 0.590 & 0.701 & \underline{0.680} & 0.535 & 0.527 & 0.607 \\
\rowcolor{blue!8} 8 & \textbf{0.614} & 0.695 & \textbf{0.714} & \textbf{0.550} & 0.540 & \underline{0.623} \\
16 & 0.579 & \textbf{0.714} & 0.637 & \underline{0.537} & 0.511 & 0.596 \\
\bottomrule
\end{tabular}
\caption{Effect of the number of task-relevant concepts $r$ on five survival prediction tasks.
Metrics are C-index under 5-fold cross-validation, and Avg.\ denotes the mean across all cohorts.
The best and second-best results are shown in \textbf{bold} and \underline{underlined}, respectively, and the highlighted row indicates the default setting used in the main paper.
}
\label{tab:r_ablation_surv}
\endgroup
\end{table}

\clearpage
\subsection{Ablation Study on the Concept-Rank Decay Factor $\alpha$}
\label{sec:alpha_ablation}
$\alpha$ controls how sharply concept importance decays with rank ($\tilde{S}_j=\alpha^{\,j-1}S_j$), where $\alpha=1$ discards the LLM-derived ranking and weights all $r$ concepts uniformly. We sweep $\alpha\in\{0.1,0.3,0.5,0.7,0.9,1.0\}$ with $r=8$, $K_1=30$, and $\beta=0.5$.

\subsubsection{Classification} 
Table~\ref{tab:alpha_ablation} shows that performance is largely flat for $\alpha \le 0.5$, with average accuracy between 80.5\% and 80.6\% and average AUC between 90.6\% and 90.7\%, and then increases for larger $\alpha$: 83.0\% / 92.0\% at $\alpha = 0.7$ and 84.1\% / 92.9\% at $\alpha = 0.9$. The difference between these two regimes is concentrated on the many-class cohorts. Relative to $\alpha = 0.9$, UBC-OCEAN loses 8.4--9.4 accuracy points and BRACS loses 7.1--9.1 accuracy points for $\alpha \le 0.5$, whereas TCGA-BRCA and TCGA-NSCLC vary by at most 1.2 and 0.9 points, respectively, across the entire sweep. TCGA-RCC attains its best result at $\alpha = 0.3$ (94.4\% / 99.1\%). Uniform weighting ($\alpha = 1.0$) is below $\alpha = 0.9$ on both averages (83.8\% / 92.6\% vs. 84.1\% / 92.9\%); it is lower on TCGA-NSCLC, TCGA-RCC, and BRCA$^{*}$, higher on TCGA-BRCA and UBC-OCEAN accuracy, and identical on BRACS. We therefore use $\alpha = 0.9$, which attains the best average on both metrics.
\begin{table*}[!htbp]
\centering
\setlength{\tabcolsep}{4.5pt}
\begin{tabular}{c cc cc cc cc cc cc cc}
\toprule
& \multicolumn{2}{c}{TCGA-BRCA} & \multicolumn{2}{c}{TCGA-NSCLC} & \multicolumn{2}{c}{TCGA-RCC} & \multicolumn{2}{c}{BRCA\textsuperscript{*}} & \multicolumn{2}{c}{UBC-OCEAN} & \multicolumn{2}{c}{BRACS} & \multicolumn{2}{c}{Avg.} \\
\cmidrule(lr){2-3}\cmidrule(lr){4-5}\cmidrule(lr){6-7}\cmidrule(lr){8-9}\cmidrule(lr){10-11}\cmidrule(lr){12-13}\cmidrule(lr){14-15}
$\alpha$ & Acc & AUC & Acc & AUC & Acc & AUC & Acc & AUC & Acc & AUC & Acc & AUC & Acc & AUC \\
\midrule
0.1             & 90.2 & \underline{93.7} & 91.6 & 96.9 & 91.6 & 97.8 & 64.7 & \textbf{82.9} & 70.7 & 88.8 & 73.9 & 83.4 & 80.5 & 90.6 \\
0.3             & \textbf{91.3} & 93.6 & 91.5 & 97.1 & \textbf{94.4} & \textbf{99.1} & 64.4 & \underline{82.1} & 70.3 & 89.0 & 71.9 & 83.6 & 80.6 & 90.8 \\
0.5             & 90.1 & \underline{93.7} & 91.2 & \textbf{97.4} & 92.1 & 98.1 & 66.2 & \underline{82.1} & 71.3 & 89.3 & 72.4 & 82.8 & 80.6 & 90.6 \\
0.7             & 90.7 & \textbf{93.8} & 91.2 & 97.0 & 92.5 & \underline{98.8} & \underline{66.9} & 81.4 & \textbf{81.2} & \underline{93.5} & \underline{75.5} & \underline{87.2} & 83.0 & 92.0 \\
\rowcolor{blue!8} 0.9             & \underline{90.9} & \textbf{93.8} & \textbf{92.1} & \underline{97.2} & \underline{93.3} & \underline{98.8} & \textbf{67.4} & 81.6 & \underline{79.7} & \textbf{94.7} & \textbf{81.0} & \textbf{91.1} & \textbf{84.1} & \textbf{92.9} \\
1.0 & \textbf{91.3} & 93.4 & \underline{91.8} & 96.8 & 91.6 & 98.6 & 66.0 & 81.2 & \textbf{81.2} & \textbf{94.7} & \textbf{81.0} & \textbf{91.1} & \underline{83.8} & \underline{92.6} \\
\bottomrule
\end{tabular}
\caption{Effect of the rank-decay factor $\alpha$ in concept importance weighting ($\tilde{S}_j=\alpha^{\,j-1}S_j$) on six classification tasks.
BRCA\textsuperscript{*} denotes molecular subtype classification on TCGA-BRCA.
Metrics are Acc\,/\,AUC (\%) under 5-fold cross-validation, and Avg.\ denotes the mean Acc and AUC across all tasks.
The best and second-best results are shown in \textbf{bold} and \underline{underlined}, respectively, and the highlighted row indicates the default setting used in the main paper.}
\label{tab:alpha_ablation}
\end{table*}

\subsubsection{Survival Prediction} 
Stronger decay is preferred: $\alpha = 0.7$ attains the best average C-index (0.623), ahead of $\alpha = 0.3$ (0.614) and $\alpha = 0.9$ (0.612). We therefore fix $\alpha$ per task family rather than tuning it per dataset.
\begin{table}[h]
\centering

\begingroup
\setlength{\tabcolsep}{2.2mm}
\begin{tabular}{@{}lcccccc@{}}
\toprule
$\alpha$ & BRCA & KIRC & KIRP & LUAD & LUSC & Avg. \\
\midrule
0.1 & 0.582 & \underline{0.711} & 0.652 & 0.508 & 0.527 & 0.596 \\
0.3 & \underline{0.587} & 0.696 & 0.690 & 0.548 & \textbf{0.550} & \underline{0.614} \\
0.5 & 0.569 & \textbf{0.716} & 0.606 & 0.519 & 0.538 & 0.590 \\
\rowcolor{blue!8} 0.7 & \textbf{0.614} & 0.695 & \underline{0.714} & \underline{0.550} & \underline{0.540} & \textbf{0.623} \\
0.9 & 0.574 & 0.695 & \textbf{0.716} & \textbf{0.557} & 0.518 & 0.612 \\
1.0 & 0.575 & 0.668 & 0.701 & 0.522 & 0.520 & 0.597 \\
\bottomrule
\end{tabular}
\caption{Effect of the rank-decay factor $\alpha$ in concept importance weighting ($\tilde{S}_j=\alpha^{\,j-1}S_j$) on five survival prediction tasks.
Metrics are C-index under 5-fold cross-validation, and Avg.\ denotes the mean C-index across all cohorts.
The best and second-best results are shown in \textbf{bold} and \underline{underlined}, respectively, and the highlighted row indicates the default setting used in the main paper.
}
\label{tab:table_alpha_ablation_survival}
\endgroup
\end{table}

\clearpage
\subsection{Ablation Study on the Standard-Deviation Weight $\beta$}
\label{sec:beta_ablation}
$\beta$ balances the two terms of the informativeness score $q=\mu+\beta\sigma$. $\beta=0$ reduces the module to mean pooling over the weighted similarity maps, whereas large $\beta$ ranks patches almost entirely by their variation across concepts. We sweep $\beta\in\{0.0,0.5,0.7,1.0,1.5,2.0\}$ with $r=8$, $K_1=30$, and $\alpha$ set to its per-task-family value.

\subsubsection{Classification}
Removing the variation term costs 1.1\% accuracy and 0.1\% AUC ($\beta=0$ vs.\ $\beta=0.5$), confirming that patches responding selectively to individual concepts carry evidence that averaging alone does not capture. Performance degrades sharply for $\beta\ge1.5$ (81.5\% accuracy at $\beta=2.0$), where the score is dominated by the variation term. $\beta=0.5$ attains the best average on both metrics, with $\beta=1.0$ the closest alternative (84.0\% accuracy).
\begin{table*}[!htbp]
\centering
\setlength{\tabcolsep}{4.5pt}
\begin{tabular}{c cc cc cc cc cc cc cc}
\toprule
& \multicolumn{2}{c}{TCGA-BRCA} & \multicolumn{2}{c}{TCGA-NSCLC} & \multicolumn{2}{c}{TCGA-RCC} & \multicolumn{2}{c}{BRCA\textsuperscript{*}} & \multicolumn{2}{c}{UBC-OCEAN} & \multicolumn{2}{c}{BRACS} & \multicolumn{2}{c}{Avg.} \\
\cmidrule(lr){2-3}\cmidrule(lr){4-5}\cmidrule(lr){6-7}\cmidrule(lr){8-9}\cmidrule(lr){10-11}\cmidrule(lr){12-13}\cmidrule(lr){14-15}
$\beta$ & Acc & AUC & Acc & AUC & Acc & AUC & Acc & AUC & Acc & AUC & Acc & AUC & Acc & AUC \\
\midrule
0.0 & 89.8 & 93.5 & 91.8 & 97.0 & \underline{92.7} & \textbf{98.8} & 64.7 & \textbf{82.8} & 80.0 & \textbf{94.9} & 79.0 & 89.7 & 83.0 & \underline{92.8} \\
\rowcolor{blue!8} 0.5 & \underline{90.9} & \textbf{93.8} & \underline{92.1} & \textbf{97.2} & \textbf{93.3} & \textbf{98.8} & \underline{67.4} & 81.6 & 79.7 & \underline{94.7} & \textbf{81.0} & \textbf{91.1} & \textbf{84.1} & \textbf{92.9} \\
0.7 & 90.7 & \underline{93.7} & \underline{92.1} & \underline{97.1} & 92.4 & \textbf{98.8} & 66.0 & 80.7 & \underline{80.6} & 94.0 & \underline{80.1} & \underline{89.9} & 83.7 & 92.4 \\
1.0 & \textbf{91.3} & 93.3 & \textbf{92.8} & \underline{97.1} & 91.4 & \underline{98.7} & \textbf{67.8} & \underline{82.4} & \textbf{80.8} & 94.0 & 79.7 & 89.8 & \underline{84.0} & 92.6 \\
1.5 & 88.2 & 89.4 & 90.7 & 96.7 & 91.9 & 98.1 & 66.4 & 81.0 & 80.0 & \underline{94.7} & 75.4 & 86.5 & 82.1 & 91.1 \\
2.0 & 87.1 & 89.0 & 91.4 & 96.7 & 92.4 & 98.1 & 65.6 & 79.8 & 78.1 & 94.4 & 74.6 & 86.3 & 81.5 & 90.7 \\
\bottomrule
\end{tabular}
\caption{Effect of the standard-deviation weight $\beta$ on six classification tasks.
BRCA\textsuperscript{*} denotes molecular subtype classification on TCGA-BRCA.
Metrics are Acc\,/\,AUC (\%) under 5-fold cross-validation, and Avg.\ denotes the mean Acc and AUC across all tasks.
The best and second-best results are shown in \textbf{bold} and \underline{underlined}, respectively, and the highlighted row indicates the default setting used in the main paper.
}
\label{tab:beta_ablation}
\end{table*}

\subsubsection{Survival Prediction}
$\beta=0.5$ attains the best average C-index (0.623). Unlike classification, the response to $\beta$ is non-monotone: performance drops at $\beta=0.7$ (0.590) before recovering to 0.603--0.607 for $\beta\ge1.0$, and no single setting is best across all cohorts. Since $\beta=0.5$ is optimal for both task families, a single value serves throughout, in contrast to $\alpha$.
\begin{table}[!htbp]
\centering

\begingroup
\setlength{\tabcolsep}{2.2mm}
\begin{tabular}{@{}lcccccc@{}}
\toprule
$\beta$ & BRCA & KIRC & KIRP & LUAD & LUSC & Avg. \\
\midrule
0.0 & 0.557 & 0.694 & \textbf{0.720} & 0.537 & 0.487 & 0.599 \\
\rowcolor{blue!8} 0.5 & 0.614 & 0.695 & \underline{0.714} & 0.550 & \textbf{0.540} & \textbf{0.623} \\
0.7 & 0.609 & \underline{0.703} & 0.605 & 0.530 & 0.505 & 0.590 \\
1.0 & \underline{0.616} & \textbf{0.708} & 0.592 & \textbf{0.568} & \underline{0.531} & 0.603 \\
1.5 & 0.611 & 0.699 & 0.659 & 0.538 & 0.525 & 0.606 \\
2.0 & \textbf{0.626} & 0.690 & 0.660 & \underline{0.557} & 0.500 & \underline{0.607} \\
\bottomrule
\end{tabular}
\caption{Effect of the standard-deviation weight $\beta$ on five survival prediction tasks.
Metrics are C-index under 5-fold cross-validation, and Avg.\ denotes the mean across all cohorts.
The best and second-best results are shown in \textbf{bold} and \underline{underlined}, respectively, and the highlighted row indicates the default setting used in the main paper.
}
\label{tab:beta_ablation_surv}
\endgroup
\end{table}

\clearpage
\subsection{Ablation Study on the Choice of VLM Backbone}
To evaluate backbone robustness, we additionally evaluate our method using an alternative pretrained VLM, PLIP~\cite{plip}, which is pretrained on expert-curated medical Twitter image–text pairs.
PLIP is used for pseudo-label assignment, CS-VPT, and coarse-to-fine patch selection, while keeping all other components—including the feature extractor—and all hyperparameters unchanged.
All methods are evaluated using the same downstream protocol and an HR patch-retention budget.\\
Table~\ref{tab:classification_plip} reports the results across six classification tasks.
LanGuSTE with PLIP achieves an average accuracy and AUC of 82.9\% and 92.2\%, respectively, outperforming all competing patch-selection methods.
Although its average performance is slightly lower than that of exhaustive 20$\times$ processing, it uses only 10\% of the HR patches retained for downstream analysis.
Table~\ref{tab:survival_plip} further shows the highest average C-index of 0.610 across the five
survival cohorts.
Although PLIP performs slightly below the default CONCH~\cite{conch} VLM backbone, these results demonstrate that LanGuSTE remains effective across different pathology-domain VLMs.
\begin{table}[!htbp]
\centering
\begingroup
\setlength{\tabcolsep}{1mm}
\small
\begin{tabular}{@{}l*{20}{c}@{}}
\toprule
\multirow{2}{*}{Method}
& \multicolumn{3}{c}{TCGA-BRCA}
& \multicolumn{3}{c}{TCGA-NSCLC}
& \multicolumn{3}{c}{TCGA-RCC}
& \multicolumn{3}{c}{BRCA\textsuperscript{*}}
& \multicolumn{3}{c}{UBC-OCEAN}
& \multicolumn{3}{c}{BRACS}
& \multicolumn{2}{c}{Avg.} \\
\cmidrule(lr){2-4}
\cmidrule(lr){5-7}
\cmidrule(lr){8-10}
\cmidrule(lr){11-13}
\cmidrule(lr){14-16}
\cmidrule(lr){17-19}
\cmidrule(l){20-21}
& Acc & AUC & $R$
& Acc & AUC & $R$
& Acc & AUC & $R$
& Acc & AUC & $R$
& Acc & AUC & $R$
& Acc & AUC & $R$
& Acc & AUC \\
\midrule
20$\times$ all
& \underline{89.6} & \textbf{93.1} & 100
& \underline{91.7} & \textbf{97.0} & 100
& 91.9 & \underline{98.4} & 100
& \underline{66.0} & \textbf{83.5} & 100
& \textbf{80.2} & \underline{94.2} & 100
& \underline{78.3} & 88.8 & 100
& \textbf{83.0} & \textbf{92.5} \\

20$\times$ 10\%
& 84.6 & 92.8 & 10
& 86.6 & 96.2 & 10
& 90.9 & 97.6 & 10
& 64.1 & 80.5 & 10
& 74.9 & 91.2 & 10
& 77.7 & 88.0 & 10
& 79.8 & 91.1 \\
\midrule
HDMIL
& \underline{89.6} & \textbf{93.1} & 10
& 91.2 & 95.8 & 10
& \underline{92.5} & 98.0 & 10
& 62.8 & 80.4 & 10
& 77.3 & 92.7 & 10
& 77.4 & 89.1 & 10
& 81.8 & 91.5 \\

FOCUS
& \textbf{89.7} & 92.7 & 10
& 91.2 & 95.8 & 10
& \textbf{92.9} & 98.0 & 10
& 64.9 & 80.9 & 10
& 76.9 & 93.1 & 10
& 78.0 & \textbf{89.9} & 10
& 82.3 & 91.7 \\

EmmPD
& 89.1 & 91.9 & 10
& 91.6 & 96.2 & 10
& 91.9 & 97.8 & 10
& 65.6 & \underline{83.1} & 10
& 76.2 & 92.8 & 10
& 74.8 & 86.3 & 10
& 81.5 & 91.4 \\

ZoomMIL
& 88.3 & 90.2 & 10
& \textbf{91.8} & 96.5 & 10
& 90.9 & 97.8 & 10
& 65.8 & 82.9 & 10
& 79.4 & 93.8 & 10
& 75.9 & 87.4 & 10
& 82.0 & 91.4 \\

HiVE-MIL
& 89.2 & \underline{93.0} & 10
& 88.8 & 95.1 & 10
& 91.0 & \textbf{98.5} & 10
& 62.3 & 78.8 & 10
& 76.5 & 93.0 & 10
& 74.8 & 87.1 & 10
& 80.4 & 90.9 \\

PATHS
& 89.3 & 91.7 & 10
& 90.6 & 96.0 & 10
& \underline{92.5} & \textbf{98.5} & 10
& 63.9 & 79.9 & 10
& 74.4 & 90.9 & 10
& 74.6 & 85.2 & 10
& 80.9 & 90.4 \\
\midrule

\textbf{Ours w/ PLIP}
& 88.3 & 91.1 & 10
& 91.4 & \underline{96.8} & 10
& 92.2 & \textbf{98.5} & 10
& \textbf{66.9} & \underline{83.1} & 10
& \underline{80.0} & \textbf{94.4} & 10
& \textbf{78.6} & \underline{89.2} & 10
& \underline{82.9} & \underline{92.2} \\
\bottomrule
\end{tabular}
\caption{Comparison of LanGuSTE using PLIP as VLM with other methods on six classification tasks under the same budget $R=10\%$.
BRCA\textsuperscript{*} denotes molecular subtype classification on TCGA-BRCA.
Metrics are Acc\,/\,AUC (\%) under 5-fold cross-validation, and Avg.\ denotes the mean Acc and AUC across all tasks.
The best and second-best results are shown in \textbf{bold} and \underline{underlined}, respectively.}
\label{tab:classification_plip}
\endgroup
\end{table}\textbf{}
\begin{table}[!htbp]
\centering

\begingroup
\small
\setlength{\tabcolsep}{2.2mm}
\begin{tabular}{@{}lcccccc@{}}
\toprule
Method & BRCA & KIRC & KIRP & LUAD & LUSC & Avg. \\
\midrule
20$\times$ all  & 0.557 & 0.656 & \underline{0.701} & 0.538 & 0.514 & 0.593 \\
20$\times$ 10\% & 0.547 & 0.676 & 0.695 & 0.530 & 0.500 & 0.590 \\
\midrule
HDMIL         & 0.589 & 0.666 & 0.627 & 0.538 & 0.512 & 0.586 \\
FOCUS         & 0.576 & 0.656 & 0.562 & 0.520 & 0.493 & 0.561 \\
EmmPD         & 0.579 & \textbf{0.693} & 0.663 & \textbf{0.559} & \underline{0.544} & \underline{0.608} \\
ZoomMIL       & 0.577 & 0.628 & \textbf{0.708} & \underline{0.551} & 0.522 & 0.597 \\
HiVE-MIL      & 0.566 & \underline{0.685} & 0.668 & \underline{0.551} & \textbf{0.571} & \underline{0.608} \\
PATHS         & \underline{0.612} & 0.659 & 0.620 & 0.533 & 0.543 & 0.593 \\
\midrule
\textbf{Ours w/ PLIP} & \textbf{0.633} & 0.682 & 0.671 & 0.544 & 0.519 & \textbf{0.610} \\
\bottomrule
\end{tabular}
\caption{Comparison of LanGuSTE using PLIP as VLM with other methods on five survival prediction tasks.
Metrics are C-index under 5-fold cross-validation, and Avg. denotes the mean across the five cohorts.
The best and second-best results are shown in \textbf{bold} and
\underline{underlined}, respectively.}
\label{tab:survival_plip}
\endgroup
\end{table}




\clearpage
\subsection{Qualitative Cross-Scale Semantic Retrieval}
\begin{figure*}[h]
\centering
\includegraphics[width=0.89\linewidth]{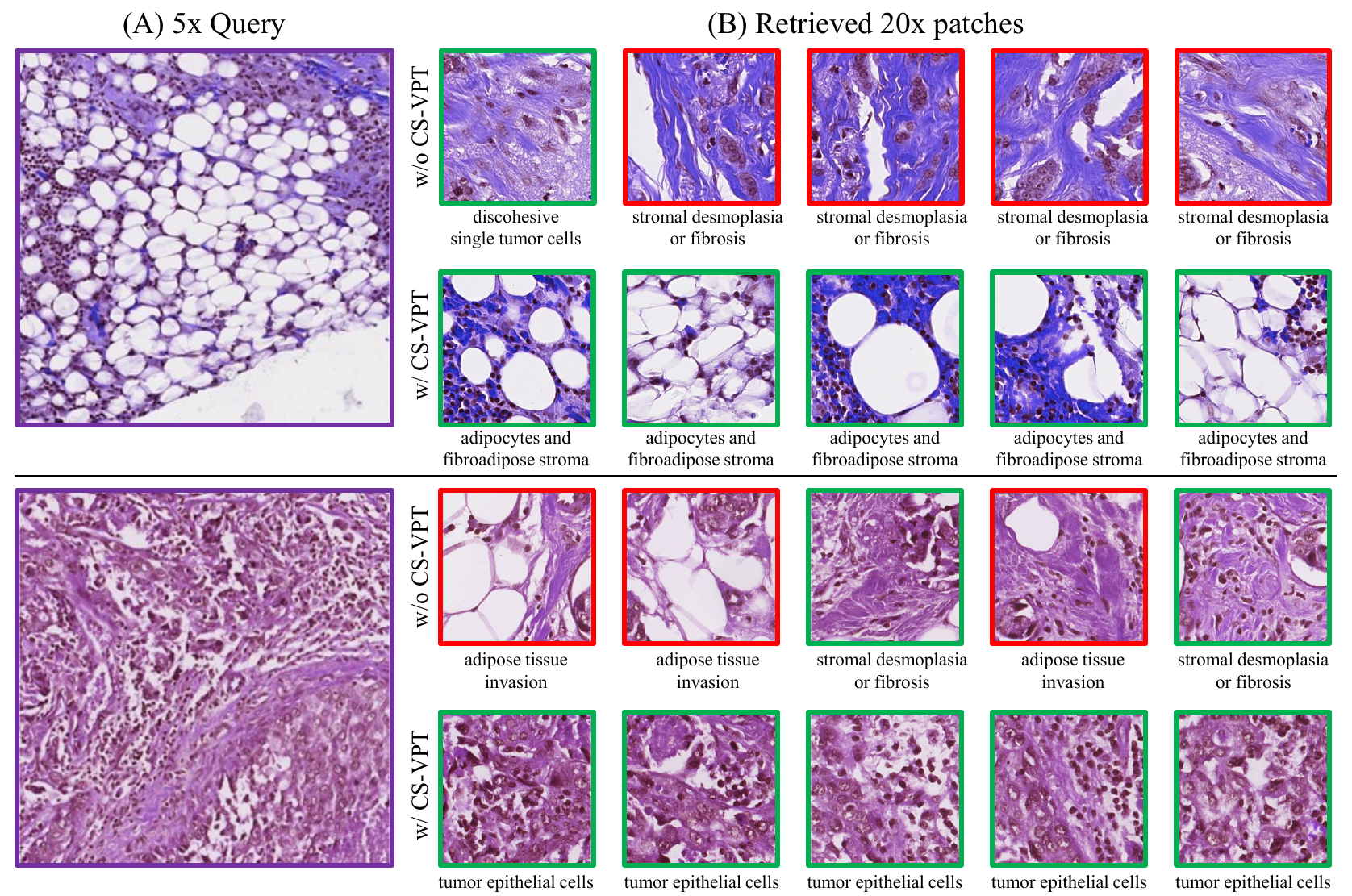}
\caption{Qualitative examples of 20$\times$ patches retrieved using 5$\times$ queries from the TCGA-BRCA cohort.}
\label{fig_supple_brca}
\end{figure*}

\begin{figure*}[h]
\centering
\includegraphics[width=0.89\linewidth]{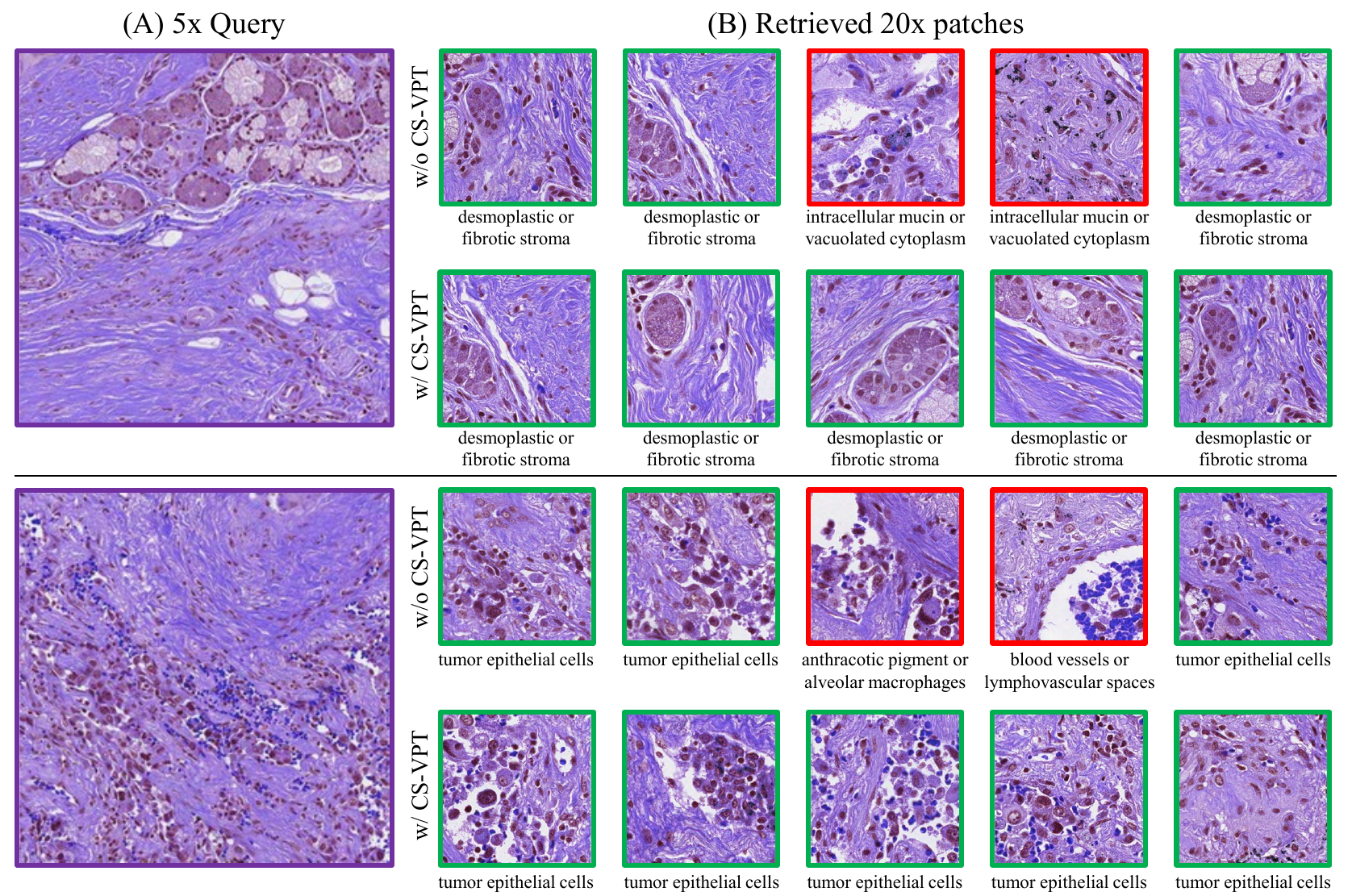}
\caption{Qualitative examples of 20$\times$ patches retrieved using 5$\times$ queries from the TCGA-NSCLC cohort.}
\label{fig_supple_nsclc}
\end{figure*}

\begin{figure*}[h]
\centering
\includegraphics[width=0.89\linewidth]{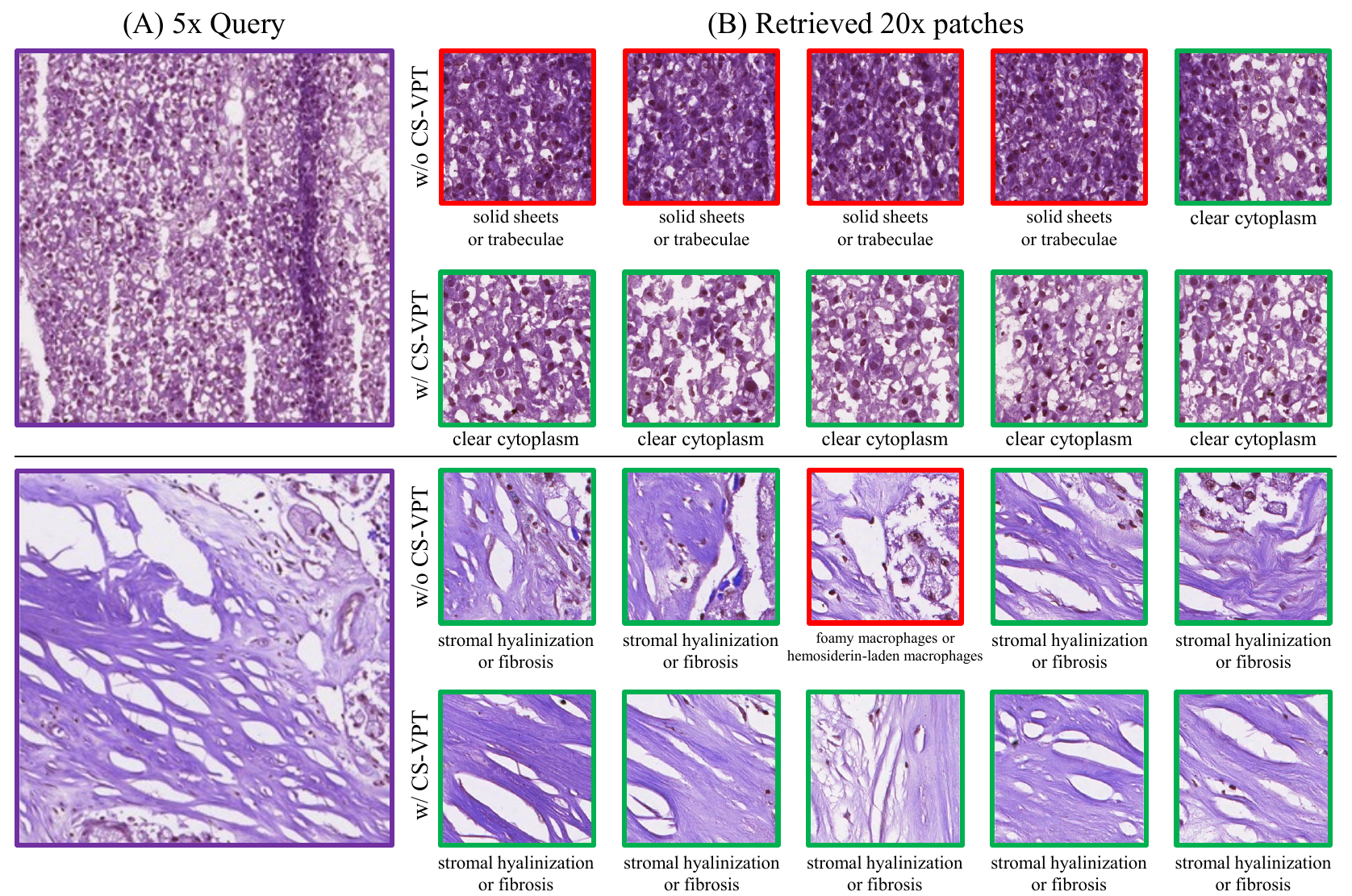}
\caption{Qualitative examples of 20$\times$ patches retrieved using 5$\times$ queries from the TCGA-RCC cohort.}
\label{fig_supple_rcc}
\end{figure*}

\begin{figure*}[h]
\centering
\includegraphics[width=0.89\linewidth]{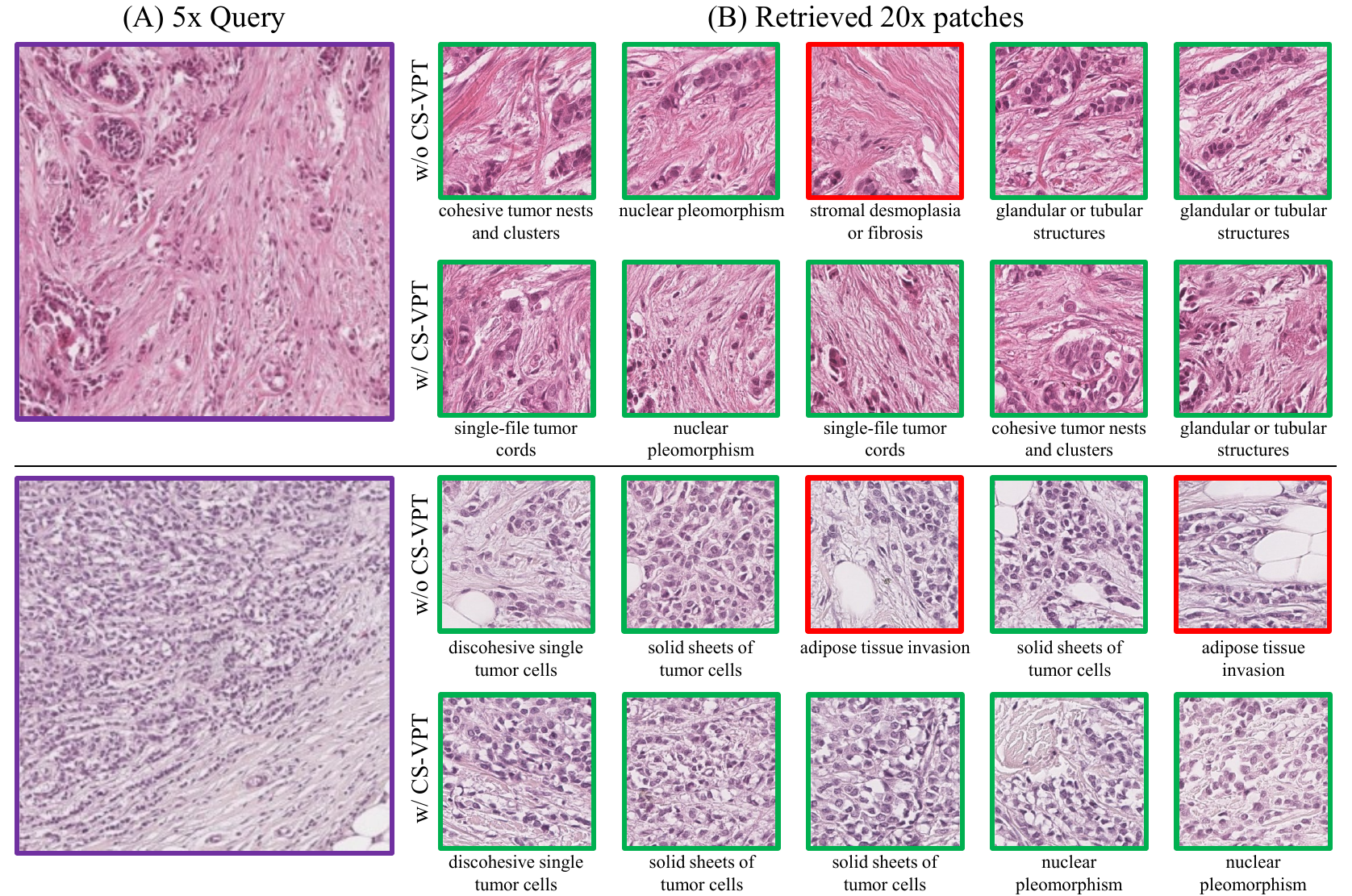}
\caption{Qualitative examples of 20$\times$ patches retrieved using 5$\times$ queries from the BRACS cohort.}
\label{fig_supple_bracs}
\end{figure*}

To qualitatively examine whether CS-VPT transfers high-resolution semantic information to low-resolution representations, we visualize cross-scale retrieval results on TCGA-BRCA (Figure~\ref{fig_supple_brca}), TCGA-NSCLC (Figure~\ref{fig_supple_nsclc}), TCGA-RCC (Figure~\ref{fig_supple_rcc}), and BRACS (Figure~\ref{fig_supple_bracs}).
For each 5$\times$ query patch, we retrieve the five most similar 20$\times$ patches from the same WSI based on cosine similarity.
The spatially corresponding child patches of the query are excluded from the retrieval candidates to prevent trivial spatial matches.
We compare retrievals obtained using the original 5$\times$ representation from the vanilla VLM and the CS-VPT-enhanced representation.
For each query, the first and second rows show the results without and with CS-VPT, respectively.
The text below each retrieved patch indicates its assigned pseudo-label.
A retrieved patch is considered semantically relevant when its pseudo-label matches that of at least one child patch of the 5$\times$ query; relevant and irrelevant retrievals are outlined in green and red, respectively.\\
Across all four cohorts, the retrieval results become more semantically consistent after applying CS-VPT.
The original 5$\times$ representations occasionally retrieve 20$\times$ patches whose pseudo-label matches one of the child labels but corresponds only to a minor visual pattern within the overall 5$\times$ query region.
For example, the first retrieved patch in the upper example of Figure~\ref{fig_supple_brca} is formally counted as relevant, yet its pseudo-label reflects only a relatively small component of the query.
In other cases, the retrieved patches are visually similar to parts of the query but are assigned pseudo-labels that are absent from the query's child-label set, as observed in the third retrieved patch of the lower examples in Figures~\ref{fig_supple_nsclc} and~\ref{fig_supple_bracs}.
These cases suggest that, without cross-scale alignment, the 5$\times$ representation may emphasize locally similar or incidental patterns rather than the dominant semantic content represented by its HR children.\\
By contrast, CS-VPT more consistently retrieves patches associated with the morphological concepts represented among the query's child patches.
This trend is observed across all datasets, and is consistent with the intended role of CS-VPT in transferring fine-grained HR semantics to LR representations.
In particular, the results provide qualitative support for the prototype-level objective: by aligning an LR patch with the prototypes of the pseudo-label clusters represented by its children, $\mathcal{L}_{\mathrm{proto}}$ appears to encourage the LR representation to capture the broader semantic characteristics shared by those HR patches, rather than merely matching isolated local appearances.

\subsection{Additional Coarse-to-Fine Selection Results}
\begin{figure*}[h]
\centering
\includegraphics[width=0.9\linewidth]{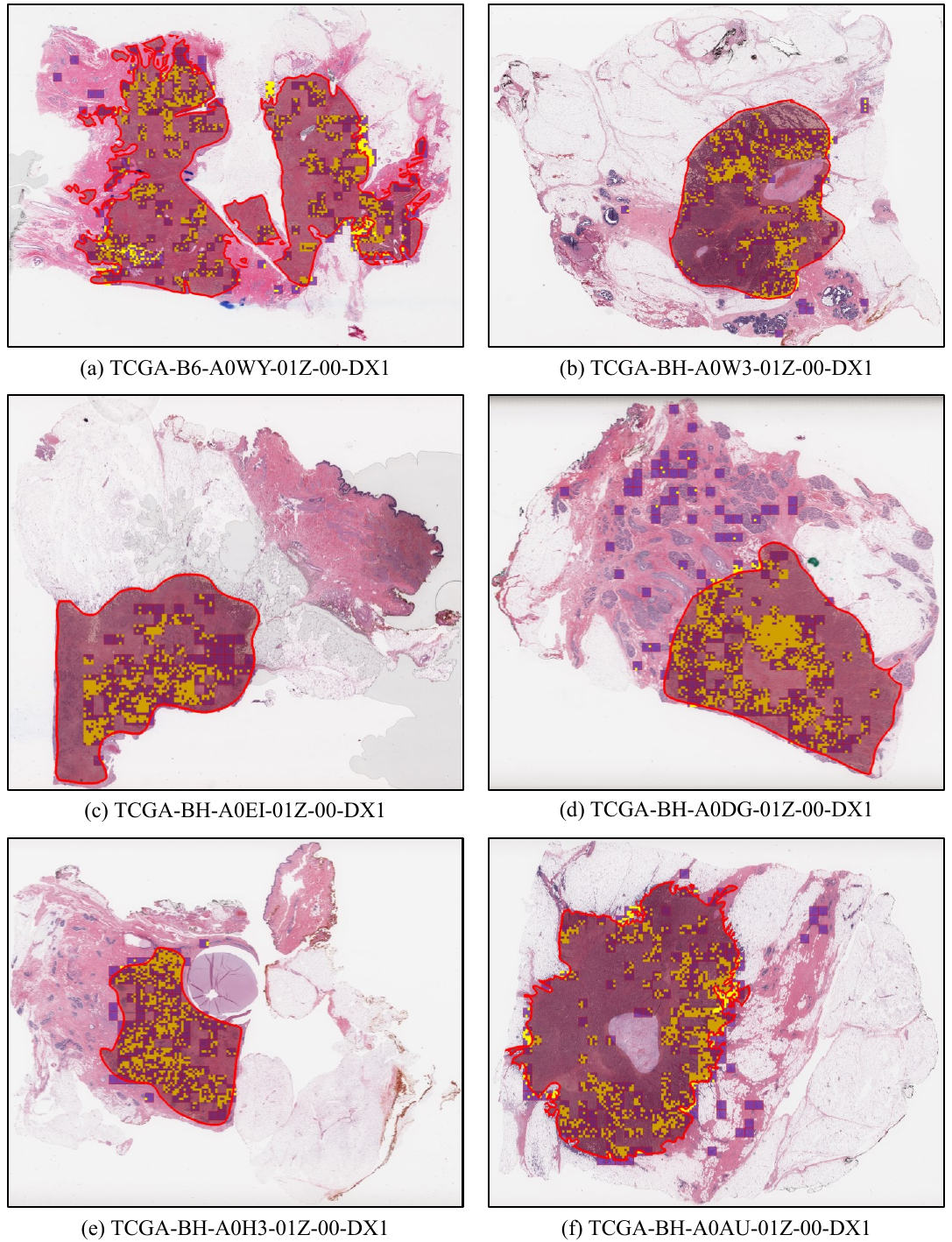}
\caption{Additional coarse-to-fine patch selection results by LanGuSTE on the TCGA-BRCA cohort. Purple boxes indicate candidate regions selected at the coarse-level stage (5$\times$); yellow boxes indicate patches selected at the fine-level stage (20$\times$) within those candidates. The red contours delineate the invasive carcinoma extent annotated by an expert pathologist. Slide identifiers are shown below each example.}
\label{fig_overlay_supple}
\end{figure*}

Figure~\ref{fig_overlay_supple} provides six additional examples of the coarse-to-fine selection results presented in the main paper, using expert annotations from TCGA-BRCA~\cite{hishi}. Across all examples, the final 20$\times$ patches are concentrated within the annotated invasive carcinoma regions, while the coarse-level candidates retained at 5$\times$ extend beyond them. This is consistent with the intended behavior of the two-stage design: the coarse stage retains a broader candidate set to avoid discarding informative regions prematurely, and the fine stage narrows the selection to patches carrying the most relevant morphological evidence. The examples also illustrate that the selection follows the shape of the annotated region: for slides with a single compact lesion, such as (b) and (d), the selected patches form one tight cluster, whereas in (a), where the annotation is split into three separate regions, the selection is likewise distributed across all of them.

\section{LLM-Derived Pathological Concepts}
Tables~\ref{tab:concepts_cla} and~\ref{tab:concepts_sur} provide the complete textual resources obtained from the LLM~\cite{gpt5} for the classification and survival prediction tasks, respectively.
For each dataset, we report the dataset- and task-specific query submitted to the LLM, together with the resulting set of 16 pathological visual features.
These features describe morphological patterns that are commonly observed in the corresponding cohort and can be identified in 20$\times$ histopathology patches.
The features in each table are numbered according to their task relevance.
Features 1--8 correspond to the top-$r$ concepts that are used to guide the language-based coarse-to-fine patch selection process, with $r=8$ in our setting.
Features 9--16 are not included in the patch-selection text pool but remain part of the full set of 16 concepts used for HR patch pseudo-labeling and cross-scale alignment in CS-VPT.

\clearpage
\twocolumn

\begin{table*}[t]
\centering

\begingroup
\small
\setlength{\tabcolsep}{1mm}

\begin{tabularx}{
    \textwidth
}{
    @{}
    >{
        \hsize=1\hsize
        \linewidth=\hsize
        \raggedright\arraybackslash
    }X
    @{\hspace{3mm}}
    >{
        \hsize=1\hsize
        \linewidth=\hsize
        \raggedright\arraybackslash
    }X
    @{}
}
\toprule[2pt]


\textbf{TCGA-BRCA}
&
\textbf{TCGA-BRCA\textsuperscript{*}}
\\
\cmidrule(r){1-1}
\cmidrule(l){2-2}

\textbf{LLM Query.}\par
\textit{%
Please list 16 most commonly observed visual features in BRCA WSI
at 20$\times$ magnification. Among them, list the top 8 that are most
important for distinguishing between IDC and ILC, in descending order
of importance.%
}
&
\textbf{LLM Query.}\par
\textit{%
Please list 16 most commonly observed visual features in BRCA WSI
at 20$\times$ magnification. Among them, list the top 8 that are most
important for distinguishing among LumA, LumB, Basal, and HER2 for
molecular subtype classification, in descending order of importance.%
}
\\
\cmidrule(r){1-1}
\cmidrule(l){2-2}

\textbf{Concept Features}\par
\noindent
\parbox[t]{0.485\linewidth}{%
    \raggedright
    \textbf{1. single-file tumor cords}\par
    \textbf{2. discohesive single \newline tumor cells}\par
    \textbf{3. targetoid or periductal growth pattern}\par
    \textbf{4. glandular or tubular structures}\par
    \textbf{5. cohesive tumor nests \newline and clusters}\par
    \textbf{6. stromal desmoplasia \newline or fibrosis}\par
    \textbf{7. adipose tissue invasion}\par
    \textbf{8. nuclear pleomorphism}%
}%
\hfill
\parbox[t]{0.485\linewidth}{%
    \raggedright
    9. solid sheets of tumor cells\par
    10. mitotic figures\par
    11. tumor necrosis\par
    12. benign mammary ducts \newline and lobules\par
    13. tumor-infiltrating lymphocytes\par
    14. blood vessels or lymphovascular spaces\par
    15. adipocytes and \newline fibroadipose stroma\par
    16. tumor epithelial cells%
}
&
\textbf{Concept Features}\par
\noindent
\parbox[t]{0.485\linewidth}{%
    \raggedright
    \textbf{1. mitotic figures}\par
    \textbf{2. nuclear pleomorphism}\par
    \textbf{3. tumor-infiltrating lymphocytes}\par
    \textbf{4. tumor necrosis}\par
    \textbf{5. glandular or \newline tubular structures}\par
    \textbf{6. solid sheets of tumor cells}\par
    \textbf{7. stromal desmoplasia \newline or fibrosis}\par
    \textbf{8. cohesive tumor nests \newline and clusters}%
}%
\hfill
\parbox[t]{0.485\linewidth}{%
    \raggedright
    9. tumor epithelial cells\par
    10. blood vessels or lymphovascular spaces\par
    11. benign mammary ducts \newline and lobules\par
    12. adipose tissue invasion\par
    13. adipocytes and \newline fibroadipose stroma\par
    14. single-file tumor cords\par
    15. discohesive single \newline tumor cells\par
    16. targetoid or periductal growth pattern%
}
\\
\midrule[2pt]


\textbf{TCGA-NSCLC}
&
\textbf{BRACS}
\\
\cmidrule(r){1-1}
\cmidrule(l){2-2}

\textbf{LLM Query.}\par
\textit{%
Please list 16 most commonly observed visual features in NSCLC WSI
at 20$\times$ magnification. Among them, list the top 8 that are most
important for distinguishing between LUAD and LUSC, in descending order
of importance.%
}
&
\textbf{LLM Query.}\par
\textit{%
Please list 16 most commonly observed visual features in BRCA WSI
at 20$\times$ magnification. Among them, list the top 8 that are most
important for distinguishing among benign, atypical, and malignant
tumors, in descending order of importance.%
}
\\
\cmidrule(r){1-1}
\cmidrule(l){2-2}

\textbf{Concept Features}\par
\noindent
\parbox[t]{0.485\linewidth}{%
    \raggedright
    \textbf{1. keratin pearls}\par
    \textbf{2. intercellular bridges}\par
    \textbf{3. acinar or glandular structures}\par
    \textbf{4. intracellular mucin or vacuolated cytoplasm}\par
    \textbf{5. lepidic growth along \newline alveolar septa}\par
    \textbf{6. squamous keratinization with dense eosinophilic cytoplasm}\par
    \textbf{7. papillary structures with fibrovascular cores}\par
    \textbf{8. micropapillary tumor clusters}%
}%
\hfill
\parbox[t]{0.485\linewidth}{%
    \raggedright
    9. solid tumor nests and sheets\par
    10. tumor necrosis\par
    11. desmoplastic or fibrotic stroma\par
    12. alveolar spaces and \newline entrapped alveolar septa\par
    13. tumor-infiltrating \newline inflammatory cells\par
    14. blood vessels or \newline lymphovascular spaces\par
    15. anthracotic pigment or \newline alveolar macrophages\par
    16. tumor epithelial cells%
}
&
\textbf{Concept Features}\par
\noindent
\parbox[t]{0.485\linewidth}{%
    \raggedright
    \textbf{1. adipose tissue invasion}\par
    \textbf{2. stromal desmoplasia \newline or fibrosis}\par
    \textbf{3. nuclear pleomorphism}\par
    \textbf{4. glandular or \newline tubular structures}\par
    \textbf{5. solid sheets of tumor cells}\par
    \textbf{6. tumor necrosis}\par
    \textbf{7. mitotic figures}\par
    \textbf{8. benign mammary \newline ducts and lobules}%
}%
\hfill
\parbox[t]{0.485\linewidth}{%
    \raggedright
    9. cohesive tumor nests \newline and clusters\par
    10. tumor epithelial cells\par
    11. blood vessels or lymphovascular spaces\par
    12. tumor-infiltrating lymphocytes\par
    13. adipocytes and \newline fibroadipose stroma\par
    14. single-file tumor cords\par
    15. discohesive single tumor cells\par
    16. targetoid or periductal growth pattern%
}
\\
\midrule[2pt]


\textbf{TCGA-RCC}
&
\textbf{UBC-OCEAN}
\\
\cmidrule(r){1-1}
\cmidrule(l){2-2}

\textbf{LLM Query.}\par
\textit{%
Please list 16 most commonly observed visual features in RCC WSI
at 20$\times$ magnification. Among them, list the top 8 that are most
important for distinguishing among CCRCC, PRCC, and CHRCC, in descending
order of importance.%
}
&
\textbf{LLM Query.}\par
\textit{%
Please list 16 most commonly observed visual features in ovarian WSI
at 20$\times$ magnification. Among them, list the top 8 that are most
important for distinguishing among CC, EC, HGSC, LGSC, and MC, in
descending order of importance.%
}
\\
\cmidrule(r){1-1}
\cmidrule(l){2-2}

\textbf{Concept Features}\par
\noindent
\parbox[t]{0.485\linewidth}{%
    \raggedright
    \textbf{1. papillary or tubulopapillary structures}\par
    \textbf{2. clear cytoplasm}\par
    \textbf{3. delicate thin-walled \newline vascular network}\par
    \textbf{4. plant-like cell borders}\par
    \textbf{5. perinuclear halos}\par
    \textbf{6. foamy macrophages or hemosiderin-laden macrophages}\par
    \textbf{7. wrinkled raisinoid nuclei}\par
    \textbf{8. psammoma bodies}%
}%
\hfill
\parbox[t]{0.485\linewidth}{%
    \raggedright
    9. fibrovascular cores\par
    10. nested or alveolar architecture\par
    11. eosinophilic granular cytoplasm\par
    12. solid sheets or trabeculae\par
    13. stromal hyalinization \newline or fibrosis\par
    14. tumor necrosis or hemorrhage\par
    15. lymphocytes or \newline inflammatory cells\par
    16. tumor epithelial cells%
}
&
\textbf{Concept Features}\par
\noindent
\parbox[t]{0.485\linewidth}{%
    \raggedright
    \textbf{1. clear cytoplasm}\par
    \textbf{2. hobnail tumor cells}\par
    \textbf{3. intracellular or \newline extracellular mucin}\par
    \textbf{4. cribriform or \newline back-to-back glands}\par
    \textbf{5. nuclear pleomorphism}\par
    \textbf{6. micropapillary or hierarchical papillary structures}\par
    \textbf{7. papillary structures with fibrovascular cores}\par
    \textbf{8. squamous morules or squamous differentiation}%
}%
\hfill
\parbox[t]{0.485\linewidth}{%
    \raggedright
    9. psammoma bodies \newline or calcifications\par
    10. cystic or tubulocystic spaces\par
    11. hyalinized stromal cores \newline or hyaline globules\par
    12. glandular or tubular structures\par
    13. solid sheets or tumor nests\par
    14. mitotic figures\par
    15. tumor necrosis\par
    16. tumor epithelial cells%
}
\\
\bottomrule[2pt]

\end{tabularx}

\endgroup

\caption{
LLM queries and corresponding ranked concept features for the six
classification tasks. TCGA-BRCA\textsuperscript{*} denotes molecular
subtype classification.
}
\label{tab:concepts_cla}

\end{table*}
\begin{table*}[t]
\centering

\begingroup
\small
\setlength{\tabcolsep}{1mm}

\begin{tabularx}{
    \textwidth
}{
    @{}
    >{
        \hsize=1\hsize
        \linewidth=\hsize
        \raggedright\arraybackslash
    }X
    @{\hspace{3mm}}
    >{
        \hsize=1\hsize
        \linewidth=\hsize
        \raggedright\arraybackslash
    }X
    @{}
}
\toprule[2pt]


\textbf{BRCA}
&
\textbf{LUAD}
\\
\cmidrule(r){1-1}
\cmidrule(l){2-2}

\textbf{LLM Query.}\par
\textit{%
Please list 16 most commonly observed visual features in BRCA WSI at 20$\times$ magnification. Among them, list the top 8 that are most important for predicting c-index-based survival risk, in descending order of importance.
}
&
\textbf{LLM Query.}\par
\textit{%
Please list 16 most commonly observed visual features in LUAD WSI at 20$\times$ magnification. Among them, list the top 8 that are most important for predicting c-index-based survival risk, in descending order of importance.
}
\\
\cmidrule(r){1-1}
\cmidrule(l){2-2}

\textbf{Concept Features}\par
\noindent
\parbox[t]{0.485\linewidth}{%
    \raggedright
    \textbf{1. mitotic figures}\par
    \textbf{2. nuclear pleomorphism}\par
    \textbf{3. glandular or tubular structures}\par
    \textbf{4. tumor necrosis}\par
    \textbf{5. tumor-infiltrating lymphocytes}\par
    \textbf{6. solid sheets of tumor cells}\par
    \textbf{7. stromal desmoplasia \newline or fibrosis}\par
    \textbf{8. blood vessels or \newline lymphovascular spaces}%
}%
\hfill
\parbox[t]{0.485\linewidth}{%
    \raggedright
    9. adipose tissue invasion\par
    10. cohesive tumor nests \newline and clusters\par
    11. tumor epithelial cells\par
    12. benign mammary ducts \newline and lobules\par
    13. adipocytes and \newline fibroadipose stroma\par
    14. single-file tumor cords\par
    15. discohesive single \newline tumor cells\par
    16. targetoid or periductal \newline growth pattern%
}
&
\textbf{Concept Features}\par
\noindent
\parbox[t]{0.485\linewidth}{%
    \raggedright
    \textbf{1. micropapillary tumor clusters}\par
    \textbf{2. solid tumor nests and sheets}\par
    \textbf{3. spread through air spaces}\par
    \textbf{4. complex glandular or \newline cribriform architecture}\par
    \textbf{5. tumor necrosis}\par
    \textbf{6. lepidic growth along \newline alveolar septa}\par
    \textbf{7. mitotic figures}\par
    \textbf{8. nuclear pleomorphism}%
}%
\hfill
\parbox[t]{0.485\linewidth}{%
    \raggedright
    9. prominent nucleoli\par
    10. desmoplastic or fibrotic stroma\par
    11. tumor-infiltrating lymphocytes\par
    12. blood vessels or \newline lymphovascular spaces\par
    13. acinar or glandular structures\par
    14. papillary structures with \newline fibrovascular cores\par
    15. intracellular mucin or \newline vacuolated cytoplasm\par
    16. tumor epithelial cells%
}
\\
\midrule[2pt]


\textbf{KIRC}
&
\textbf{LUSC}
\\
\cmidrule(r){1-1}
\cmidrule(l){2-2}

\textbf{LLM Query.}\par
\textit{%
Please list 16 most commonly observed visual features in KIRC WSI at 20$\times$ magnification. Among them, list the top 8 that are most important for predicting c-index-based survival risk, in descending order of importance.
}
&
\textbf{LLM Query.}\par
\textit{%
Please list 16 most commonly observed visual features in LUSC WSI at 20$\times$ magnification. Among them, list the top 8 that are most important for predicting c-index-based survival risk, in descending order of importance.
}
\\
\cmidrule(r){1-1}
\cmidrule(l){2-2}

\textbf{Concept Features}\par
\noindent
\parbox[t]{0.485\linewidth}{%
    \raggedright
    \textbf{1. sarcomatoid or rhabdoid differentiation}\par
    \textbf{2. tumor necrosis}\par
    \textbf{3. vascular or \newline lymphovascular invasion}\par
    \textbf{4. prominent nucleoli}\par
    \textbf{5. nuclear pleomorphism}\par
    \textbf{6. mitotic figures}\par
    \textbf{7. solid sheets or compact tumor nests}\par
    \textbf{8. blood-filled sinusoidal spaces or hemorrhage}%
}%
\hfill
\parbox[t]{0.485\linewidth}{%
    \raggedright
    9. eosinophilic granular cytoplasm\par
    10. stromal hyalinization \newline or fibrosis\par
    11. trabecular or acinar architecture\par
    12. nested or alveolar architecture\par
    13. clear cytoplasm\par
    14. delicate thin-walled \newline vascular network\par
    15. cystic change or microcysts\par
    16. tumor epithelial cells%
}
&
\textbf{Concept Features}\par
\noindent
\parbox[t]{0.485\linewidth}{%
    \raggedright
    \textbf{1. tumor budding or \newline single-cell invasion}\par
    \textbf{2. mitotic figures}\par
    \textbf{3. nuclear pleomorphism}\par
    \textbf{4. tumor necrosis}\par
    \textbf{5. basaloid tumor areas}\par
    \textbf{6. solid tumor nests and sheets}\par
    \textbf{7. squamous keratinization with dense eosinophilic cytoplasm}\par
    \textbf{8. tumor-infiltrating lymphocytes}%
}%
\hfill
\parbox[t]{0.485\linewidth}{%
    \raggedright
    9. infiltrative tumor margins\par
    10. desmoplastic or fibrotic stroma\par
    11. blood vessels or \newline lymphovascular spaces\par
    12. prominent nucleoli\par
    13. keratin pearls\par
    14. intercellular bridges\par
    15. adjacent bronchial epithelium \newline or alveolar spaces\par
    16. tumor epithelial cells%
}
\\
\midrule[2pt]


\textbf{KIRP}
&
\\
\cmidrule(r){1-1}

\textbf{LLM Query.}\par
\textit{%
Please list 16 most commonly observed visual features in KIRP WSI at 20$\times$ magnification. Among them, list the top 8 that are most important for predicting c-index-based survival risk, in descending order of importance.
}
&
\\
\cmidrule(r){1-1}

\textbf{Concept Features}\par
\noindent
\parbox[t]{0.485\linewidth}{%
    \raggedright
    \textbf{1. vascular or \newline lymphovascular invasion}\par
    \textbf{2. tumor necrosis or hemorrhage}\par
    \textbf{3. prominent nucleoli}\par
    \textbf{4. nuclear pleomorphism}\par
    \textbf{5. mitotic figures}\par
    \textbf{6. solid sheets or compact tumor nests}\par
    \textbf{7. eosinophilic granular cytoplasm}\par
    \textbf{8. papillary or tubulopapillary structures}%
}%
\hfill
\parbox[t]{0.485\linewidth}{%
    \raggedright
    9. stromal edema, hyalinization, \newline or fibrosis\par
    10. clear cytoplasmic change\par
    11. tubular or acinar structures\par
    12. pale or basophilic cytoplasm\par
    13. foamy macrophages or \newline hemosiderin-laden macrophages\par
    14. psammoma bodies \newline or calcifications\par
    15. fibrovascular cores\par
    16. tumor epithelial cells%
}
&
\\
\bottomrule[2pt]

\end{tabularx}

\endgroup

\caption{
LLM queries and corresponding ranked concept features for the five survival prediction tasks.
}
\label{tab:concepts_sur}

\end{table*}

\end{document}